\documentclass[journal,transmag]{IEEEtran}
\usepackage{cite}

\usepackage{graphicx}
\ifCLASSINFOpdf
\else
\fi
\usepackage{amsmath}
\usepackage{amssymb}
\usepackage{gensymb}
\usepackage{textcomp}

\usepackage{enumitem}
\usepackage{booktabs}
\usepackage{multirow}

\usepackage{caption}
\usepackage{subcaption}

\usepackage[pagebackref=true,breaklinks=true,letterpaper=true,colorlinks=false,bookmarks=false]{hyperref}

\DeclareMathOperator*{\argmin}{arg\,min}
\newcommand\InsertFailure[1]{
    \hspace{-12pt}
    \begin{subfigure}[c]{0.19\linewidth}
        \centering
        \includegraphics[trim={1.2cm 0 0 0},clip,width=\textwidth]{./Figures/supp/failures/#1-size.pdf}
    \end{subfigure}
    ~
    \begin{subfigure}[c]{0.19\linewidth}
        \centering
        \includegraphics[width=\textwidth]{./Figures/supp/failures/#1-pred-1-enc.jpg}
    \end{subfigure}
    \begin{subfigure}[c]{0.19\linewidth}
        \centering
        \includegraphics[width=\textwidth]{./Figures/supp/failures/#1-pred-2-enc.jpg}
    \end{subfigure}
    ~~~
    \begin{subfigure}[c]{0.19\linewidth}
        \centering
        \includegraphics[width=\textwidth]{./Figures/supp/failures/#1-min-1-enc.jpg}
    \end{subfigure}
    \begin{subfigure}[c]{0.19\linewidth}
        \centering
        \includegraphics[width=\textwidth]{./Figures/supp/failures/#1-min-2-enc.jpg}
    \end{subfigure}
    \hspace{-12pt}
}

\newcommand\InsertQualitative[1]{
    \hspace{-12pt}
    \begin{subfigure}[c]{0.19\linewidth}
        \centering
        \includegraphics[trim={1.2cm 0 0 0},clip,width=\textwidth]{./Figures/supp/qualitative/#1-size.pdf}
    \end{subfigure}
    ~
    \begin{subfigure}[c]{0.19\linewidth}
        \centering
        \includegraphics[width=\textwidth]{./Figures/supp/qualitative/#1-pred-1-enc.jpg}
    \end{subfigure}
    \begin{subfigure}[c]{0.19\linewidth}
        \centering
        \includegraphics[width=\textwidth]{./Figures/supp/qualitative/#1-pred-2-enc.jpg}
    \end{subfigure}
    ~~~
    \begin{subfigure}[c]{0.19\linewidth}
        \centering
        \includegraphics[width=\textwidth]{./Figures/supp/qualitative/#1-center-1-enc.jpg}
    \end{subfigure}
    \begin{subfigure}[c]{0.19\linewidth}
        \centering
        \includegraphics[width=\textwidth]{./Figures/supp/qualitative/#1-center-2-enc.jpg}
    \end{subfigure}
    \hspace{-12pt}
}

\begin{document}
%
\title{Learning Compressible 360\textdegree~Video Isomers}
%
%
%

\author{
\IEEEauthorblockN{Yu-Chuan~Su and
Kristen~Grauman}
\IEEEauthorblockA{Department of Computer Science, The University of Texas at Austin}
}

\maketitle

\begin{abstract}
Standard video encoders developed for conventional narrow field-of-view video are widely applied to $360$\textdegree~video as well, with reasonable results.
However, while this approach commits arbitrarily to a projection of the spherical frames, we observe that some orientations of a $360$\textdegree~video, once projected, are more compressible than others.
We introduce an approach to predict the sphere rotation that will yield the maximal compression rate.
Given video clips in their original encoding, a convolutional neural network learns the association between a clip's visual content and its compressibility at different rotations of a cubemap projection.
Given a novel video,
our learning-based approach efficiently infers the most compressible direction in one shot,
without repeated rendering and compression of the source video.
We validate our idea on thousands of video clips and multiple popular video codecs.
The results show that this untapped dimension of $360$\textdegree~compression has substantial potential---``good'' rotations are typically $8{-}10\%$ more compressible than bad ones,
and our learning approach can predict them reliably $82\%$ of the time.
\end{abstract}


\section{Introduction}

Both the technology and popularity of
$360\degree$ video has grown rapidly in recent years, for emerging Virtual Reality (VR) applications and others.
Sales of $360\degree$ cameras are expected to grow by $1500\%$ from 2016 to 2022~\cite{360camera}.
Foreseeing the tremendous opportunities in $360\degree$ video,
many companies are investing in it.  
For example, Facebook and YouTube have offered $360\degree$ content support since 2015.
Facebook users have since uploaded more than one million $360\degree$ videos~\cite{fb360videostatistics},
and YouTube plans to bring $360\degree$ videos to even broader platforms (TV, gaming consoles).
$360\degree$ editing tools are now available in popular video editors such as PowerDirector and Premiere Pro.
Meanwhile, on the research side, there is strong interest in improving $360\degree$ video display~\cite{kasahara2015first,kopf2016tog,kamali2011stabilizing,su2016accv,su2017cvpr,hu2017deep,lai2017semantic},
and performing visual processing efficiently on the new format~\cite{khasanova2017graph,cohen2017convolutional,su2017nips}.
All together, these efforts make $360\degree$ video production and distribution easier and more prevalent than ever.

\begin{figure*}[t]
    \center
    \includegraphics[width=\linewidth]{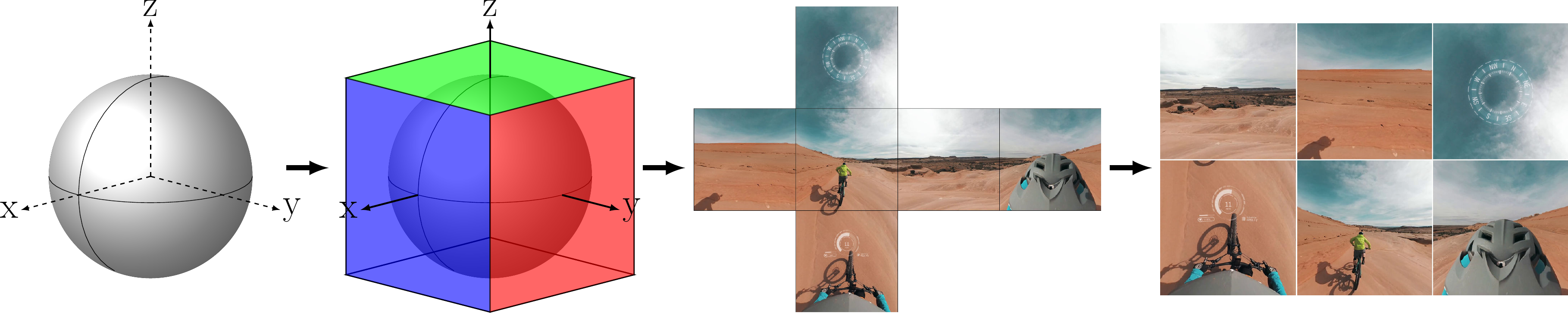}
    \caption{
        Cubemap format transformation.
        The $360\degree$ video is first projected to a cube enclosing the unit sphere and then unwrapped into 6 faces.
        The 6 faces are re-arranged to form a rectangular picture to fit video compression standards ($2{\times}3$ frame on the right).
    }
    \label{fig:faces}
\end{figure*}

At the core of all video technologies is the data format.
In particular, a compressed video bit-stream format is the basis for all video related applications,
ranging from video capture, storage, processing to distribution.
Without adequate compression,
all of the above suffer.  
$360\degree$ video is no exception.
Thus far, the focus for $360\degree$ video compression is to find a proper projection that transforms a $360\degree$ frame into a rectangular planar image that will have a high compression rate.
A current favorite is to project the sphere to a \emph{cubemap} and unwrap the cube into a planar image~\cite{fb2015cubemap,google2017eac,mpeg120} (see Fig.~\ref{fig:faces}).
Cubemaps can improve the compression rate by up to $25\%$ compared to the previously popular equirectangular projection~\cite{fb2016compressionrate}.

\begin{figure}[t]
    \center
    \includegraphics[width=1.\linewidth]{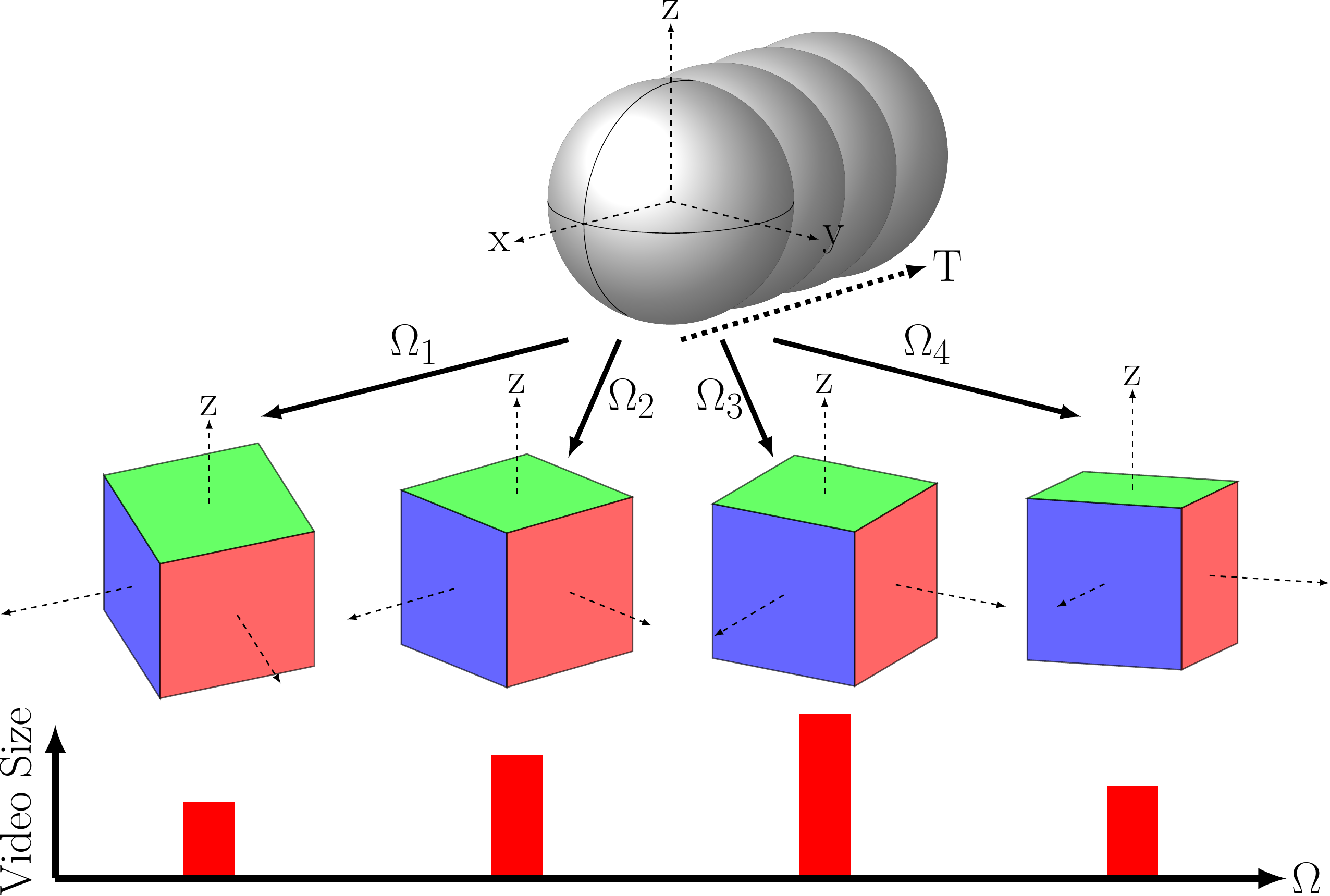}
    \caption{
        Our approach learns to automatically rotate the $360\degree$ video axis before storing the video in cubemap format.
        While the $360\degree$ videos are equivalent under rotation (``isomers''),
        the bit-streams are not because of the video compression procedures.  Our approach analyzes the video's visual content to predict its most compressible isomer.
        \label{fig:approach}
    }
\end{figure}

One unique property of $360\degree$ video is that each spherical video has an \emph{infinite number of equivalents related by a rotation.}
Therefore,
each $360\degree$ video could be transformed into multiple possible cubemaps by changing the orientation of the cube, yet all of them represent the very same video content.
We refer to these content-equivalent rotations as \emph{$360\degree$ isomers}.\footnote{Strictly speaking isomers are equivalent only theoretically, because pixels are discretely sampled and rotating a cubemap requires interpolating the pixels.  Nevertheless, as long as the pixel density, i.e.~video resolution, is high enough, the information delta is negligible.}
The isomers, however, are \emph{not} equivalents in terms of compression.  Different isomers interact differently with a given compression algorithm and so yield different compression rates (See Fig.~\ref{fig:approach}).
This is because the unwrapped cubemap is not a homogenous perspective image.
Therefore, some of the properties that current compression algorithms exploit in perspective images do not hold.
For example, while the content is smooth and continuous in perspective images,
this need not be true along an inter-face boundary in an unwrapped cubemap.
The discontinuity can introduce artificial high frequency signals and large abrupt motions,
both of which harm the compression rate (cf.~Sec.~\ref{sub:data_analysis} and Fig.~\ref{fig:concept}).
In short, our key insight is that the compression rate of a $360\degree$ video will depend on the orientation of the cubemap it is projected on.

We propose a learning-based approach to predict---from the video's visual content itself---the cubemap orientation that will minimize the video size.
First we demonstrate empirically that the orientation of a cubemap does influence the compression rate,
and the difference is not an artifact of a specific encoder but a general property over a variety of popular video formats.
Based on that observation,
we propose to automatically re-orient the cubemap for every group of pictures (GOP).\footnote{a collection of successive pictures within a coded video stream.}
A naive solution would enumerate each possible orientation, compress the GOP, and pick the one with the lowest encoded bit-stream size.
However, doing so would incur substantial overhead during compression, prohibitively costly for many settings.
Instead, our approach renders the GOP for a \emph{single} orientation after predicting the optimal orientation from the video clip rendered in its canonical orientation.
Given encoded videos in a fixed orientation,
we train a Convolutional Neural Network (CNN) that takes both the 
segmentation contours and motion vectors
in the encoded bit-stream and predicts the orientation that will yield the minimum video size.
By avoiding rendering and encoding the video clip in all possible orientations,
our approach greatly reduces the computational cost and strikes a balance between speed and compression rate.

The key benefit of our approach is a higher compression rate for $360\degree$ video that requires only to re-render the cubemap.  In particular, our idea does not require changing the video format nor the compression algorithm,
which makes it fully compatible with any existing video codec.
This is especially important in the realm of video compression,
because a new video format often takes years to standardize and deploy,
and so changing the bit-stream format would incur very high overhead.
The only additional information that our method needs to encode is the selected orientation of each GOP,
which can easily be encoded as meta data (and may become part of the standard in the future~\cite{omaf2017wd}).

We evaluate our approach on 7,436 clips containing varying content.
We demonstrate our idea has consistent impact across three popular encoders,
with video size reductions up to 76\% and typical reductions of about 8\%.
Across all videos, our learning approach achieves on average 82\% of the best potential compression rate available for all feasible isomers.

\section{Related Work}

\paragraph{360$\degree$ video analysis}
\label{par:_360degree_video}
Recent research explores ways to improve the user experience of watching $360\degree$ videos,
including stabilizing the videos~\cite{kasahara2015first,kopf2016tog,kamali2011stabilizing} or directing the field-of-view (FOV) automatically~\cite{su2016accv,su2017cvpr,hu2017deep,lai2017semantic}.
Other works study visual features in $360\degree$ images such as detecting SIFT~\cite{hansen2007scale} or learning a CNN either from scratch~\cite{khasanova2017graph,cohen2017convolutional} or from an existing model trained on ordinary perspective images~\cite{su2017nips}.
All of these methods offer new applications of $360\degree$ videos, and they assume the inputs are in some given form, e.g., equirectangular projection.
In contrast, we address learning to optimize the data format of $360\degree$ video, which can benefit many applications.

\paragraph{360$\degree$ video compression}
$360\degree$ video has sparked initial interest in new video compression techniques.
A Call for Evidence this year for a meeting on video standards~\cite{cfe2017JVET} calls attention to the need for compression techniques specific to $360\degree$ video,
and responses indicate that substantial improvement can be achieved in test cases~\cite{JVET-G0023,JVET-G0024,JVET-G0025,JVET-G0026}.
Whereas these efforts aim for the next generation in video compression standards,
 our method is compatible with existing video formats and can be applied directly without any modification of existing video codecs.
For video streaming, some work studies the value in devoting more bits to the region of $360\degree$ content currently viewed by the user~\cite{sanchez2015panohevc,sreedhar2016adaptive}.

\paragraph{Projection of spherical images}
\label{par:video_format}

$360\degree$ image projection has long been studied in the field of map projection.
As famously proven by Gauss, no single projection can project a sphere to a plane without introducing some kind of distortion.
Therefore, many different projections are proposed,
each designed to preserve certain properties such as distance, area, direction, etc.~\cite{snyder1987map}.
For example, the popular equirectangular projection preserves the distance along longitude circles.
Various projection models have been developed to improve perceived quality for $360\degree$ images.
Prior work~\cite{zelnik2005squaring} studies how to select or combine the projections for a better display, and others develop new projection methods to minimize visual artifacts~\cite{kim-iccv2017,chang-iccv2013}.
Our work is not about the human-perceived quality of a projected $360\degree$ image; rather, the mode of projection is relevant to our problem only in regards to how well the resulting stack of 2D frames can be compressed.

Cubemap is adopted as one of the two presentations for $360\degree$ video in the MPEG Omnidirectional MediA Format (OMAF)~\cite{mpeg120},
which is likely to become part of future standards,
and major $360\degree$ video sharing sites such as YouTube and Facebook have turned to the new format~\cite{fb2015cubemap,google2017eac}.
Cubemaps can improve the compression rate by $25\%$ compared to equirectangular projection, which suffers from redundant pixels and distorted motions~\cite{fb2016compressionrate}.
The Rotated Sphere Projection is an alternative to cubemap projection with fewer discontinuous boundaries~\cite{adeel2017rsp}.
Motivated by the compression findings~\cite{fb2016compressionrate}, our approach is built upon the standard cubemap format.
Our method is compatible with existing data formats and can offer a further reduction of video size at almost zero cost.

\paragraph{Deep learning for image compression}
\label{par:deep_learning_for_compression}

Recent work investigates ways to improve image compression using deep neural networks.
One common approach is to improve predictive coding using either a feed-forward CNN~\cite{santurkar2017generative,rippel2017real} or recurrent neural network (RNN)~\cite{toderici2015variable,toderici2016full,johnston2017improved}.
The concept can also be extended to video compression~\cite{santurkar2017generative}.
Another approach is to allocate the bit rate dynamically using a CNN~\cite{li2017learning}.
While we also study video compression using a CNN,
we are the first to study $360\degree$ video compression, and---CNN or otherwise---the first to exploit spherical video orientation to improve compression rates.
Our idea is orthogonal to existing video compression algorithms, which could be combined with our approach without any modification to further improve performance.

\section{Cubemap Orientation Analysis}
\label{sec:analysis}

Our goal is to develop a computationally efficient method that exploits a cubemap's orientation for better compression rates.
In this section, we perform a detailed analysis on the correlation between encoded video size and cubemap orientation.
The intent is to verify that orientation is indeed important for $360\degree$ video compression.
We then introduce our method to utilize this correlation in Sec.~\ref{sec:approach}.

First we briefly review fundamental video compression concepts, which will help in understanding where our idea has leverage.  
Modern video compression standards divide a video into a series of groups of pictures (GOPs), 
which can be decoded independently to allow fast seeking and error recovery.
Each GOP starts with an \emph{I-frame},
or intra-coded picture,
which is encoded independently of other frames like a static image.
Other frames are encoded as inter-coded pictures,
and are divided into rectangular blocks.
The encoder finds a reference block in previous frames for each block that minimizes their difference.
Instead of encoding the pixels directly,
the encoder encodes the relative location of the reference block,
i.e.,~the \emph{motion vector},
and the residual between the current and reference block.
This inter-frame prediction allows encoders to exploit temporal redundancy in the video.
Note that the encoder has the freedom to fall back to intra-coding mode for blocks in an inter-coded frame if no reference block is found.

Just like static image compression, 
the encoder performs transform coding by transforming the pixels in I-frames and residuals in inter-coded frames into the frequency domain and encoding the coefficients.
The transformation improves the compression rate because high frequency signals are usually few in natural images,
and many coefficients will be zero.
To further reduce the video size,
video compression formats also exploit spatial redundancy through intra-prediction,
which predicts values to be encoded using adjacent values that are previously encoded.
The encoder will encode only the residual between the prediction and real value.
This applies to both the motion vector and transformed coefficients encoding.
Most of the residuals will be small and can be encoded efficiently using entropy coding.
For a more complete survey, see~\cite{videocompression}.

\subsection{Data Preparation}
\label{sub:data_collection}

To study the correlation between cubemap orientation and compression rate,
we collect a $360\degree$ video dataset from YouTube.
Existing datasets~\cite{su2016accv,hu2017deep} contain videos with arbitrary quality, many with 
compression artifacts that could bias the result.
Instead, we collect only high quality videos using the 4K filter in YouTube search.
We use the keyword ``360 video'' together with the $360\degree$ filter to search for videos and manually filter out those consisting of static images or CG videos.
The dataset covers a variety of video content and recording situations,
including but not limited to aerial, underwater, sports, animal, news, and event videos,
and the camera can be either static or moving.
We download the videos in equirectangular projection with 3,840 pixels width encoded in H264 high profile.

\begin{table}[t]
    \small
    \tabcolsep=0.12cm
    \center
    \begin{tabular}{llccc}
    \toprule
                                   &        & H264            & HEVC           & VP9\\
    \midrule
        \multirow{2}{*}{Video $r$ (\%)} & Avg.   & $8.43 \pm 2.43$ & $8.11 \pm 2.03$ & $7.83 \pm 2.34$\\
                                   & Range  & [4.34, 15.18]   & [4.58, 13.67]   & [3.80, 14.72]\\
    \midrule
        \multirow{2}{*}{Clip $r$ (\%)}  & Avg.   & $10.37 \pm 8.79$& $8.88 \pm 8.23$ & $9.78 \pm 8.62$\\
                                   & Range  & [1.08, 76.93]   & [1.40, 74.95]   & [1.70, 75.84]\\
    \bottomrule
    \end{tabular}
    \caption{
        Achievable video size reduction through rotation for each of three encoders.
        We can reduce the video size by up to $76\%$ by optimally changing the cubemap orientation.
    }
    \label{tab:encoders}
\end{table}

We next transcode the video into cubemap format and extract the video size in different orientations.
Because it is impossible to enumerate all possible cubemap orientations over time,
we discretize the problem by dividing the video into 2 second clips and encode each clip independently.
This is compliant with the closed GOP structure,
except that video codecs usually have the flexibility to adjust the GOP length within a given range.
For example, the default x264 encoder limits the GOP length between 25-250 frames,
i.e.~roughly 1-10 seconds,
and a common constraint for Blu-ray videos is 1-2 seconds~\cite{bluray}.
This results in a dataset consisting of 7,436 video clips from 80 videos with 4.2 hours total length.

For each clip, we sample the cubemap orientation
\begin{equation}
    \Omega = (\phi, \theta) \in \Phi \times \Theta
\end{equation}
with different yaw ($\phi$) and pitch ($\theta$) in
 $\Theta = \Phi = \{-45\degree, -40\degree, \cdots, 45\degree\}$,
i.e.,~every $5\degree$ between $[-45\degree,~45\degree]$.
This yields $|\Phi \times \Theta| = 361$ different orientations.
We restrict the orientation within $90\degree$ because of the rotational symmetry along each axis.

For each orientation,
we transform the video into cubemap format using the transform360 filter\footnote{\url{https://github.com/facebook/transform360}} in FFMPEG released by Facebook with 960 pixels resolution for each face.
Fig.~\ref{fig:faces} illustrates the transformation.
The video is then encoded using off-the-shelf encoders.
We encode the video into three popular formats---H264 using x264\footnote{\url{https://www.videolan.org/developers/x264.html}},
HEVC using x265\footnote{\url{http://x265.org}},
and VP9 using libvpx\footnote{\url{https://chromium.googlesource.com/webm/libvpx/}}.
Among them, H264 is currently the most common video format.
HEVC, also known as H265, is the successor of H264 and is the latest video compression standard.
VP9 is a competitor of HEVC developed by Google and is most popular in web applications.
We use lossless compression for all three formats to ensure rotational symmetry and extract the size of the final encoded bit-stream.
See supp.~for the exact encoding parameters.
Note that we use popular open source tools for both cubemap rendering and video compression to ensure that they are well optimized and tested.  This way any size changes we observe can be taken as common in $360\degree$ video production instead of an artifact of our implementation.

\subsection{Data Analysis}
\label{sub:data_analysis}

\begin{figure}[t]
    \center
    \includegraphics[width=1.\linewidth]{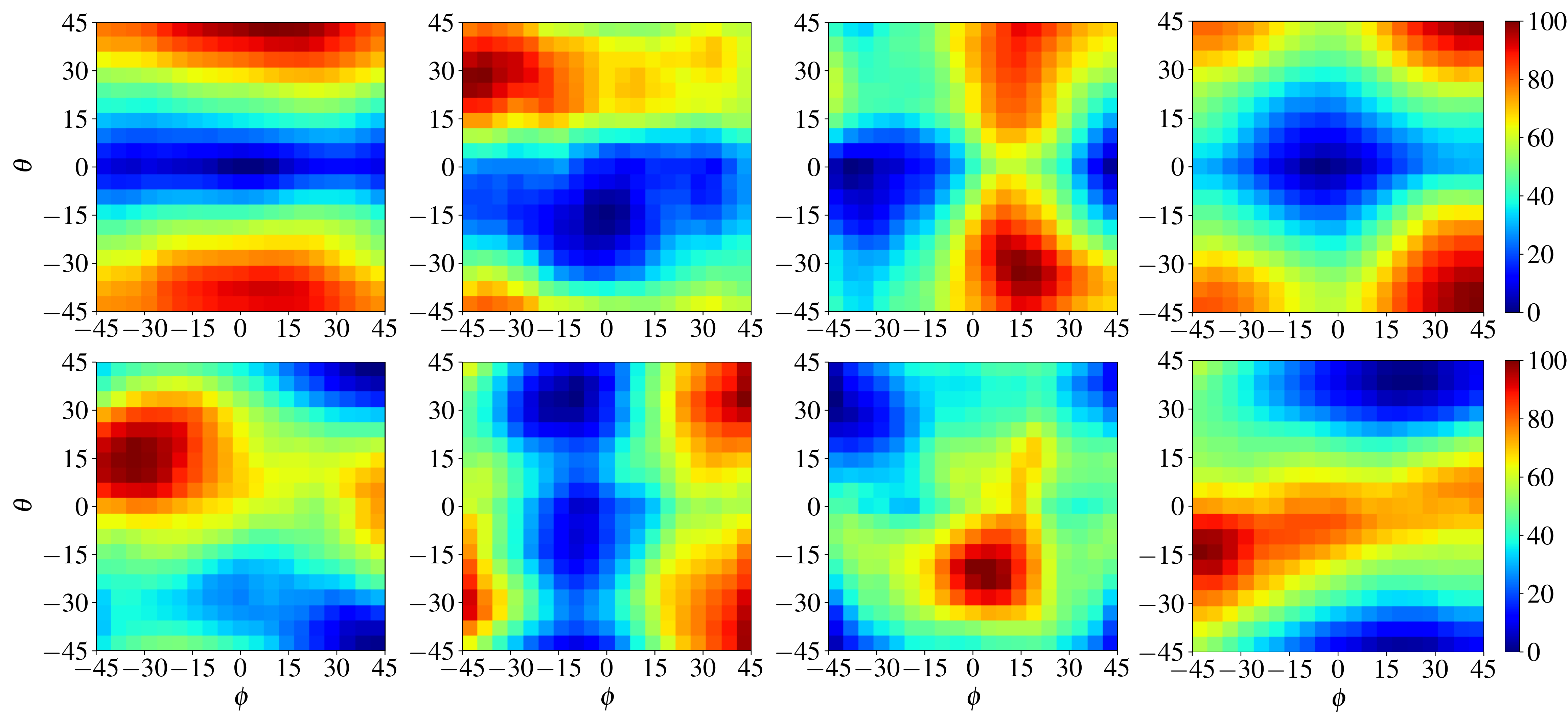}
    \caption{
        \label{fig:size_distribution}
        Relative clip size distribution w.r.t. $\Omega$.
        We cluster the distribution into 16 clusters and show 8 of them.
    }
\end{figure}

\begin{figure}[t]
    \centering
    \includegraphics[trim={1cm 0 0 0},clip,width=\linewidth]{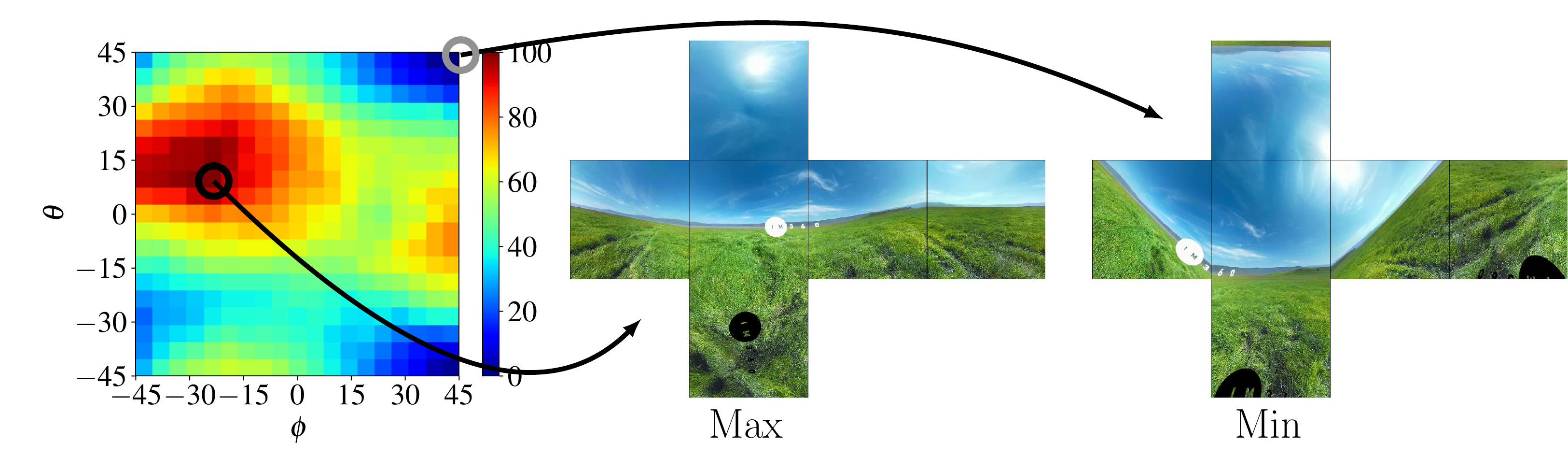}
    \caption{
        Clip size distribution of a single clip.
        We also show the cubemaps corresponding to $\Omega_{max}/\Omega_{min}$.
    }
    \label{fig:example_cube}
\end{figure}

Next we investigate how much and why the orientation of an isomer matters for compressibility.
If not mentioned specifically, all the results are obtained from H264 fromat.

\paragraph{Achievable video size reduction}
We first examine the size reduction we can achieve by changing the cubemap orientation.
In particular, we compute the \emph{reduction}
\begin{equation}
    r = 100 \times \frac{S_{\Omega^{max}}-S_{\Omega^{min}}}{S_{\Omega^{max}}},
\end{equation}
where $S_{\Omega}$ is the encoded bit-stream size with orientation $\Omega$ and $\Omega^{max}$/$\Omega^{min}$ corresponds to the orientation with maximum/minimum bit-stream size.

\begin{figure*}[t]
    \center
    \begin{subfigure}[c]{0.305\linewidth}
        \centering
        \includegraphics[width=\textwidth]{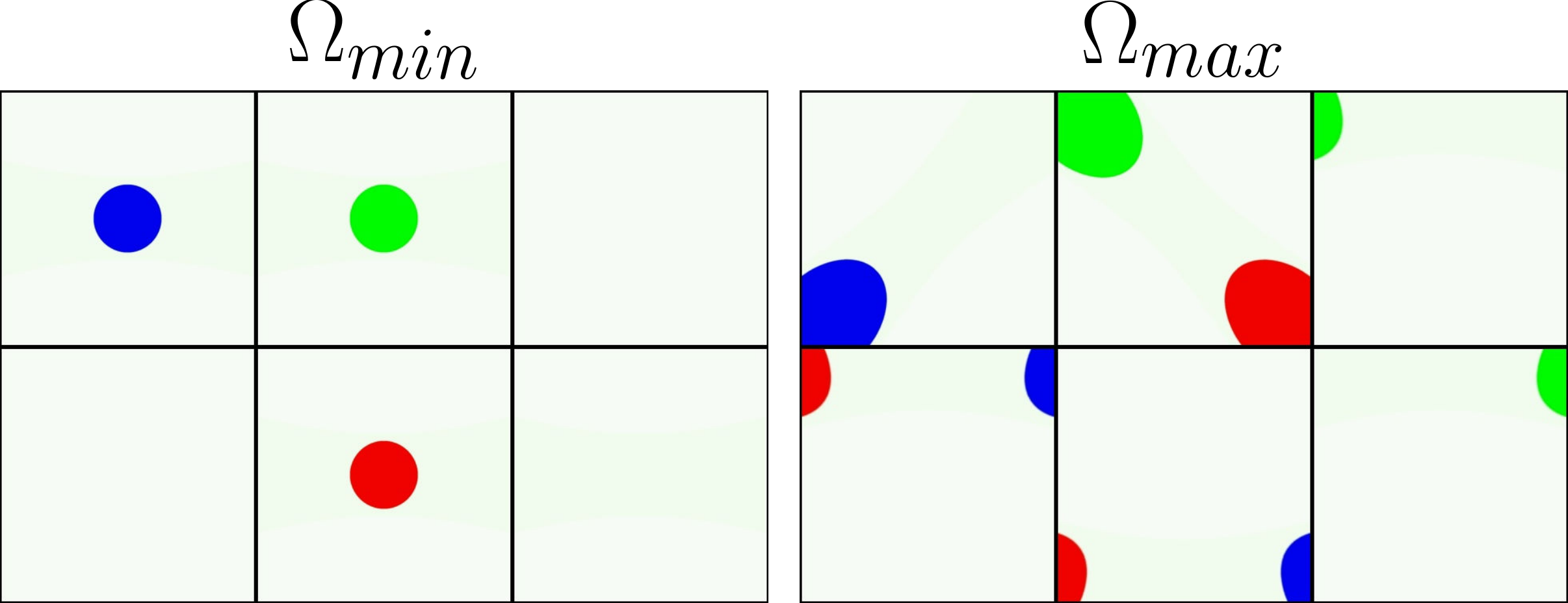}
        \caption{
            Content discontinuity.
        }
    \end{subfigure}
    ~~~
    \begin{subfigure}[c]{0.65\linewidth}
        \centering
        \includegraphics[width=\textwidth]{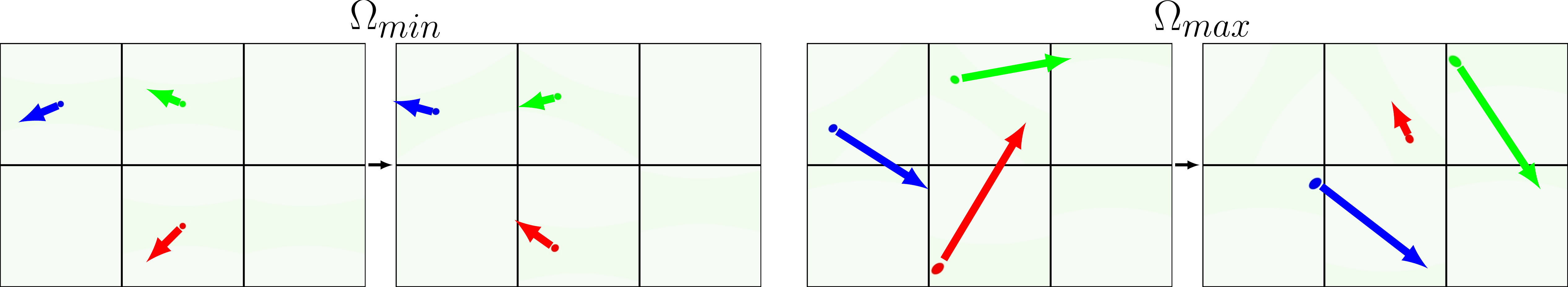}
        \caption{
            Motion discontinuity.
        }
    \end{subfigure}
    \caption{
        \label{fig:concept}
        Explanations for why different $\Omega$ have different compression rate, shown for good ($\Omega_{min}$) and bad ($\Omega_{max}$) rotations.
        (a) From a static picture perspective,
        some $\Omega$ introduce content discontinuity and reduce spatial redundancy.
        (b) From a dynamic picture perspective,
        some $\Omega$ make the motion more disordered and break the temporal redundancy.
    }
\end{figure*}

\begin{figure*}[t]
    \centering
    \includegraphics[trim={1cm 0 0 0},clip,width=\linewidth]{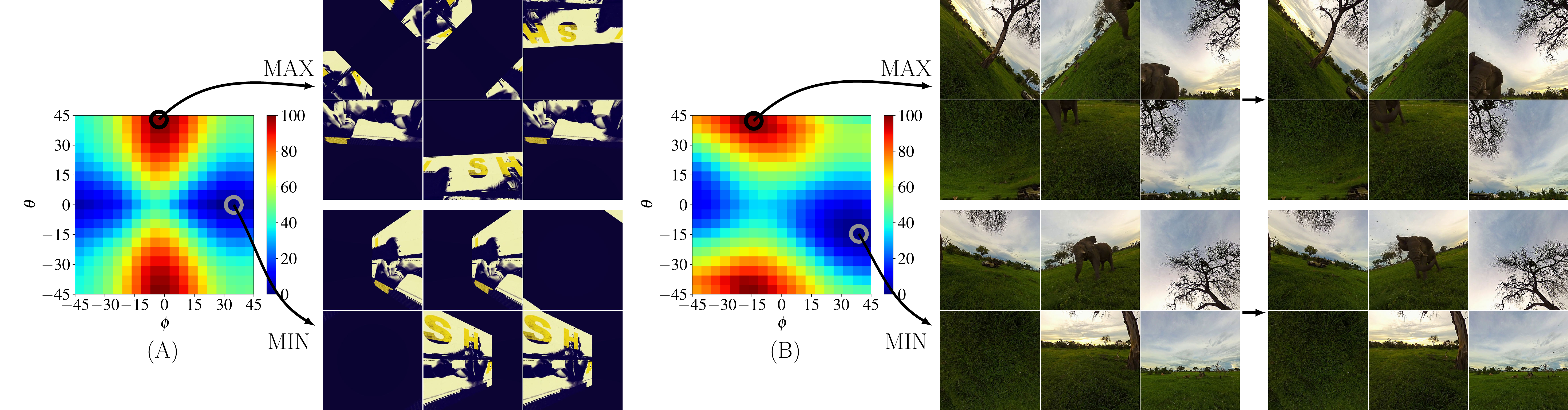}
    \caption{
        \label{fig:concept_example}
        Real examples for the explanations in Fig.~\ref{fig:concept}.
        Example (A) shows content discontinuity introduced by rotation.
        Example (B) shows motion discontinuity.
        The encoder fails to find reference blocks in this example,
        and the number of intra-coded blocks increases.
    }
\end{figure*}

Table~\ref{tab:encoders} shows the results.  
For example, the average video size reduction $\overline{r}$ is $8.43\%$ for H264,
which means that we can reduce the overall $360\degree$ video size by more than $8\%$ through rotating the video axis.
This corresponds to a 2GB reduction in our 80 video database and would scale to 25.3TB for a million video database.
The range of $r$ for each clip is $[1.08, 76.93]$,
which indicates that the compression rate is strongly content dependent,
and the size reduction can be up to $77\%$ for a single video if we allow the encoder to re-orient the $360\degree$ video.
If we restrict the rotation to $\phi$ and fix $\theta=0\degree$,
$\overline{r}$ will drop to $2.35\%$.
This result suggests that it is important to allow rotation along both axes.  
Finally we see that the average and range of reductions is quite similar across encoders, indicating that compressibility of isomers is not unique to a particular codec.

\paragraph{Video size distribution w.r.t. $\Omega$}

We next show the video size distribution with respect to $\Omega$.
We compute the \emph{normalized clip size}
\begin{equation}
    \tilde{S}_{\Omega} = 100 \times \frac{S_{\Omega} - S_{\Omega^{min}}}{S_{\Omega^{max}}-S_{\Omega^{min}}}
    \label{eq:normalized_size}
\end{equation}
for every $\Omega$ and cluster the size distribution of each clip using K-Means.
Each cluster is represented by the nearest neighbor to the center.

Fig.~\ref{fig:size_distribution} shows the results.
We can see $\Omega^{min}$ lies on or near $\theta{=}0\degree$ in half the clusters.
In general, this corresponds to orienting the cubemap perpendicular to the ground such that the top face captures the sky and the bottom face captures the camera and ground.
See Fig.~\ref{fig:faces} for example.
The top and bottom faces tend to have smaller motion within the faces in these orientations,
and the compression rate is higher because the problem reduces from compressing six dynamic pictures to four dynamic pictures plus two near static pictures.  
However, $\theta{=}0\degree$ is not best for every clip, and there are multiple modes visible in Fig.~\ref{fig:size_distribution}.
For example, the minimum size occurs at $\theta{=}\phi{=}45\degree$ in Fig.~\ref{fig:example_cube}.
Therefore, again we see it is important to allow two-dimensional rotations.

\paragraph{Reasons for the compression rate difference}

Why does the video size depend on $\Omega$?  
The fundamental reason is that all the video compression formats are designed for perspective images and heavily exploit the image properties.
The unwrapped cubemap format is a perspective image only locally within each of the six faces.
The cubemap projection introduces perspective distortion near the face boundaries and artificial discontinuities across face boundaries,
both of which make the cubemap significantly different from perspective images and can degrade the compression rate.
Because the degradation is content dependent,
different orientations result in different compression rates.

More specifically,
the reasons for the compression rate difference can be divided into two parts.
From the static image perspective,
artificial edges may be introduced if continuous patterns fall on the face boundary.
See Fig.~\ref{fig:concept} (a) and Fig.~\ref{fig:concept_example} for examples.
The edges introduce additional high frequency signals and reduce the efficiency of transform coding.
Furthermore,
the single continuous patch is divided into multiple patches that are dispersed to multiple locations in the image.
This reduces the spatial redundancy and breaks the intra-prediction.

From the dynamic video perspective,
the face boundaries can introduce abrupt jumps in the motion.
If an object moves across the boundary,
it may be teleported to a distant location on the image.
See Fig.~\ref{fig:concept} (b) and Fig.~\ref{fig:concept_example} for examples.
The abrupt motion makes it difficult to find the reference block during encoding,
and the encoder may fall back to intra-coding mode which is much less efficient.
Even if the encoder successfully finds the reference block,
the motion vectors would have very different magnitude and direction compared to those within the faces,
which breaks intra-prediction.
Finally,
because the perspective distortion is location dependent,
the same pattern will be distorted differently when it falls on different faces,
and the residual of inter-frame prediction may increase.
The analysis applies similarly across the three formats, which makes sense, since their compression strategies are broadly similar.

\paragraph{Video size correlation across formats}

Next we verify the correlation between video size and orientation is not an artifact of the specific video format or encoder.
We first compare the size reduction that can be achieved through rotation using different encoders (Table \ref{tab:encoders}).  
We can clearly see that the dependency between the compression rate and $\Omega$ is not the consequence of a specific video encoder.
Instead, it is a common property across current video compression formats.
Although different encoders have different compression rate improvements,
the differences are relatively minor,
which indicates that the problem cannot be solved by simply using a more advanced video compression standard designed for ordinary perspective images.

We further test the correlation between video size of different encoders.
We first compute the relative size of every orientation as
\begin{equation}
    S^{\prime}_{\Omega} = S_{\Omega} - S_{0,0},
    \label{eq:relative_size}
\end{equation}
where $S_{0,0}$ denotes the non-rotated source video, and then compute the correlation between encoders for each $\Omega$.
We report the average correlation across all $\Omega$ in Table~\ref{tab:correlations}.
The high correlation again verifies that the correlation between $\Omega$ and video size is common across video formats.

\begin{table}[t]
    \small
    \center
    \begin{tabular}{lccc}
    \toprule
    Encoders & H264 / H265 & H264 / VP9 & H265 / VP9\\
    \midrule
        Avg. $\rho$ & 0.8757  & 0.9533 & 0.8423 \\
    \bottomrule
    \end{tabular}
    \caption{
        The correlation of relative video sizes across video formats.
        The high correlation indicates that the dependency between video size and $\Omega$ is common across formats.
    }
    \label{tab:correlations}
\end{table}

\section{Approach}
\label{sec:approach}

\begin{figure*}[t]
    \center
    \includegraphics[width=1.\linewidth]{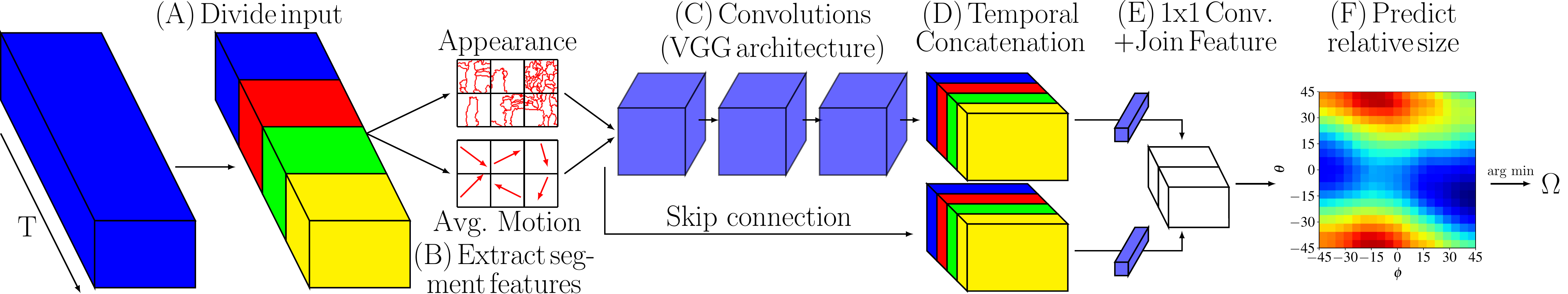}
    \caption{
        Our model takes a video clip as input and predicts $\Omega^{min}$ as output.
        (A) It first divides the video into 4 segments temporally and (B) extracts appearance and motion features from each segment.
        (C) It then concatenates the appearance and motion feature maps and feeds them into a CNN.
        (D) The model concatenates the outputs of each segment together and joins the output with the input feature map using skip connections to form the video feature.
        (F) It then learns a regression model that predicts the relative video size $S^{\prime}_{\Omega}$ for all $\Omega$ and takes the minimum one as the predicted optimally compressible isomer.
    }
    \label{fig:model}
\end{figure*}

In this section,
we introduce our approach for improving $360\degree$ video compression rates by predicting the most compressible isomer.
Given a $360\degree$ video clip,
our goal is to identify $\Omega^{min}$ to minimize the video size.
A naive solution is to render and compress the video for all possible $\Omega$ and compare their sizes.
While this guarantees the optimal solution,
it introduces a significant computational overhead, i.e.,~360 times more computation than encoding the video with a fixed $\Omega$.
For example,
it takes more than 15 seconds to encode one single clip using the default x264 encoder on a 48 core machine with Intel Xeon E5-2697 processor,
which corresponds to $15s \times 360 \approx 1.5$ hours for one clip if we try to enumerate $\Omega$.
Moreover, the computational cost will grow quadratically if we allow more fine-grained control.
Therefore, enumerating $\Omega$ is not practical.

Instead, we propose to predict $\Omega^{min}$ from the raw input without rerendering the video.
Given the input video in cubemap format,
we extract both motion and appearance features (details below) and feed them into a CNN that predicts the video size $S_{\Omega}$ for each $\Omega$,
and the final prediction of the model is
\begin{equation}
    \Omega^{min} = \argmin_{\Omega} S_{\Omega}.
\end{equation}
See Fig.~\ref{fig:model}.
The computational cost remains roughly the same as transcoding the video because the prediction takes less than a second,
which is orders of magnitude shorter than encoding the video and thus negligible.  
Since no predictor will generalize perfectly, there is a chance of decreasing the compression rate in some cases.
However, experimental results show that it yields very good results and strikes a balance between computation time and video size.

Because our goal is to find $\Omega^{min}$ for a given video clip,
exact prediction of $S_{\Omega}$ is not necessary.
Instead,
the model predicts the relative video size $S^{\prime}_{\Omega}$ from Eq.~\ref{eq:relative_size}.
The value $S^{\prime}_{\Omega}$ is scaled to $[0, 100]$ over the entire dataset to facilitate training.
We treat it as a regression problem and learn a model that predicts $361$ real values using L2 loss as the objective function.
Note that we do not predict $S_{\Omega}$ in Eq.~\ref{eq:normalized_size} because it would amplify the loss for clips with smaller size,
which may be harmful for the absolute size reduction.

We first divide the input video into 4 equal length segments.
For each segment,
we extract the appearance and motion features for each frame and average them over the segment.
For appearance features,
we segment the frame into regions using SLIC~\cite{achanta2012slic} and take the segmentation contour map as feature.
The segmentation contour represents edges in the frame,
which imply object boundaries and high frequency signals that take more bits in video compression.

For motion features,
we take the motion vectors directly from the input video stream encoding, as opposed to computing optical flow. 
The motion vectors are readily available in the input and thus this saves computation.
Furthermore, motion vectors provide more direct information about the encoder.
Specifically,
we sample one motion vector every 8 pixels and take both the forward and backward motion vectors as the feature.
Because each motion vector consists of both spatial and temporal displacement,
this results in a 6-dimensional feature.
For regions without a motion vector, we simply pad 0 for the input regardless of the encoding mode.
We concatenate the appearance and motion feature to construct a feature map with depth 7.
Because the motion feature map has lower resolution than the video frame,
we downscale the appearance feature map by 8 to match the spatial resolution.
The input resolution of each face of the cube map is therefore $960/8=160$ pixels.

The feature maps for each segment are then fed into a CNN and concatenated together as the video feature.
We use the VGG architecture~\cite{simonyan2014very} except that we increase the number of input channels in the first convolution layer.
Because fine details are important in video compression,
we use skip connections to combine low level information with high level features, following models for image segmentation~\cite{long2015fully}.
In particular,
we combine the input feature map and final convolution output as the segment feature after performing 1x1 convolution to reduce the dimension to 4 and 64 respectively.
The video feature is then fed into a fully-connected layer with $361$ outputs as the regression model.
Note that we remove the fully-connected layers in the VGG architecture to keep the spatial resolution for the regression model and reduce model size.

Aside from predicting $S_{\Omega}$, in preliminary research 
we also tried other objective functions such as regression for $\Omega^{min}$ directly or predicting $\Omega^{min}$ from the 361 possible $\Omega$ with $361$-way classification,
but none of them perform as well as the proposed approach.
Regressing $\Omega^{min}$ often falls back to predicting $(\theta, \phi) = (0, 0)$ because the distribution is symmetric.
Treating the problem as $361$-way classification has very poor accuracy because the number of training data is small and imbalanced.
We also examined 3D convolution instead of explicitly feeding the motion information as input,
but we find that 3D convolution is hard to train and performs worse than 2D convolution.

\section{Experiments}
\label{sec:experiments}

To evaluate our method,
we compute the size reduction it achieves on the $360\degree$ video dataset introduced in Sec.~\ref{sec:analysis}.

\paragraph{Baselines}
\label{par:baselines}

Because we are the first to study how to predict the cubemap orientation for better compression,
we compare our method with the following two heuristics:
\begin{itemize}[leftmargin=*,label=$\bullet$]
    \item \textsc{Random} --- Randomly rotate the cubemap to one of the 361 orientations.
        This represents the compression rate when we have no knowledge about the video orientation.
    \item \textsc{Center} --- Use the orientation provided by the videographer.
        This is a strong prior, usually corresponding to the direction of the videographer's gaze or movement and lying on the horizon of the world coordinate.
\end{itemize}

\paragraph{Evaluation metrics}
\label{par:evaluation_metrics}

We compare each method using the normalized size reduction $\tilde{r} = 1 - \tilde{S}$ for each video.
Specifically, we compute the largest full-video size by choosing $\Omega^{max}$ for every clip and sum the clip sizes.
Similarly, we compute the minimum video size.
Given the predicted orientation for each clip,
we compute the video size by rotating the cubemap by the predicted orientation.
The result indicates the fraction of reduction the method achieves compared to the optimal result.

To train and test the model,
we divide the dataset into 4 folds, each containing 20 videos.
Three are used for training, and the other is used for testing.
We report the average result over 4 folds as the final performance.

\paragraph{Implementation details}
\label{par:implementation_details}

We initialize the weights using an ImageNet pre-trained VGG model provided by the authors~\cite{simonyan2014very}.
For the first layer, we replicate the weights of the original network to increase the number of input channels.
Weights that are not in the original model are randomly initialized using Xavier initialization~\cite{glorot2010initialization}.
We train the model using ADAM~\cite{kingma2014adam} for 4,000 iterations with batch size 64 parallelized to 16 GPUs.
The base learning rate is initialized to $1.0 \times 10^{-3}$ and is decreased by a factor of 10 after 2,000 iterations.
We also apply $L_{2}$ regularization with the weight set to $5.0 \times 10^{-4}$ and use dropout for the fully-connected layers with ratio 0.5.
For SLIC,
we segment each face of the cubemap independently into 256 superpixels with compactness $m{=}1$.
The low compactness value leads to more emphasis on the color proximity in the superpixels.

\subsection{Results}
\label{sub:results}

We first examine the size reduction our method achieves.
Table~\ref{tab:size_reduction} shows the results.  
Our method performs better than the baselines in all video compression formats by $7\%-16\%$.
The improvement over the baseline is largest in HEVC,
which indicates that the advantage of our approach will become more significant as HEVC gradually replaces H264.
Interestingly,
the \textsc{Center} baseline performs particularly worse in HEVC.
The reason is that HEVC allows the encoder to achieve good compression rates in more diversed situations,
so the distribution of $\Omega^{min}$ becomes more dispersed.
The result further shows the value in considering cubemap orientation during compression as more advanced video codecs are used.
While there remains a $20\%$ room for improvement compared to the optimal result (as ascertained by enumerating $\Omega$),
our approach is significantly faster and takes less than $0.3\%$ the computation.

\begin{table}[t]
    \small
    \center
    \begin{tabular}{lccc}
    \toprule
    & H264 & HEVC & VP9\\
    \midrule
    \textsc{Random} & 50.75 & 51.62 & 51.20 \\
    \textsc{Center} & 74.35 & 63.34 & 72.92 \\
    \midrule
    \textsc{Ours}   & 82.10 & 79.10 & 81.55\\
    \bottomrule
    \end{tabular}
    \caption{
        Size reduction of each method.
        The range is $[0, 100]$, the higher the better.
    }
    \label{tab:size_reduction}
\end{table}

\begin{figure}[t]
    \center
    \begin{subfigure}[c]{0.31\linewidth}
        \centering
        \includegraphics[trim={1.2cm 0 0 0},clip,width=\textwidth]{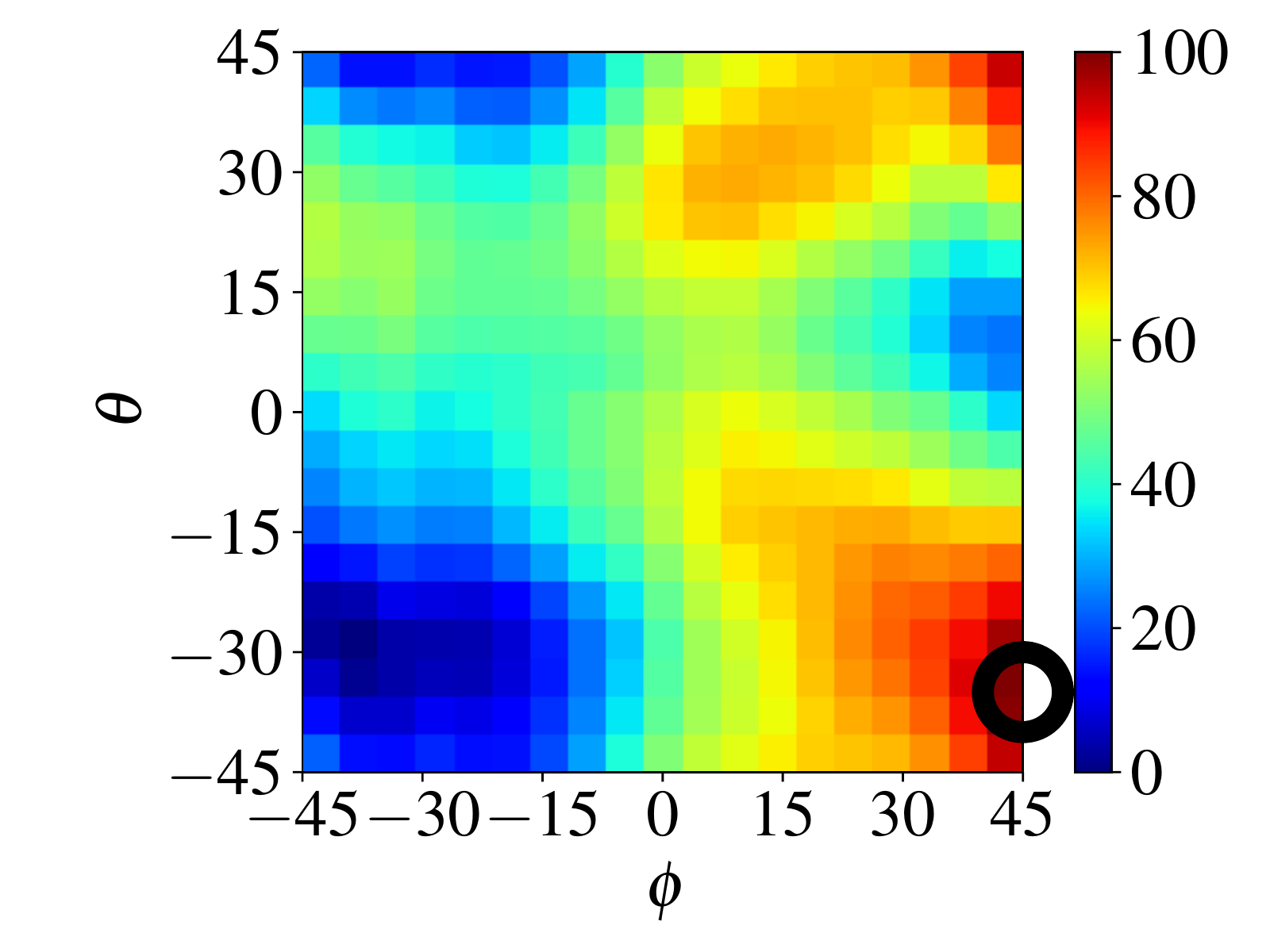}
    \end{subfigure}
    \begin{subfigure}[c]{0.33\linewidth}
        \centering
        \includegraphics[width=\textwidth]{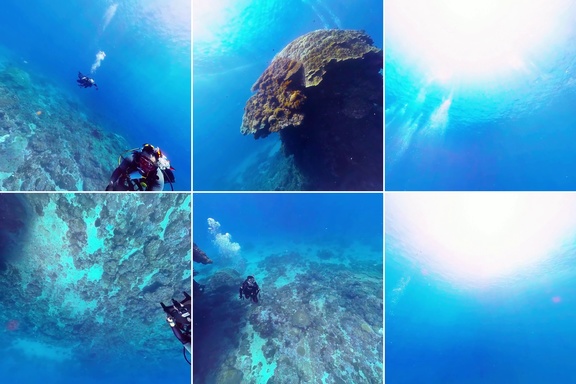}
    \end{subfigure}
    \begin{subfigure}[c]{0.33\linewidth}
        \centering
        \includegraphics[width=\textwidth]{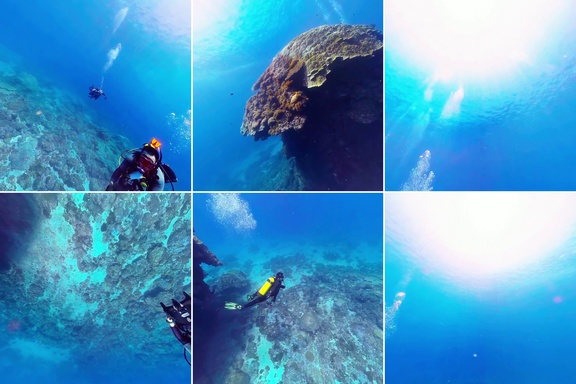}
    \end{subfigure}
    \\
    \begin{subfigure}[c]{0.31\linewidth}
        \centering
        \includegraphics[trim={1.2cm 0 0 0},clip,width=\textwidth]{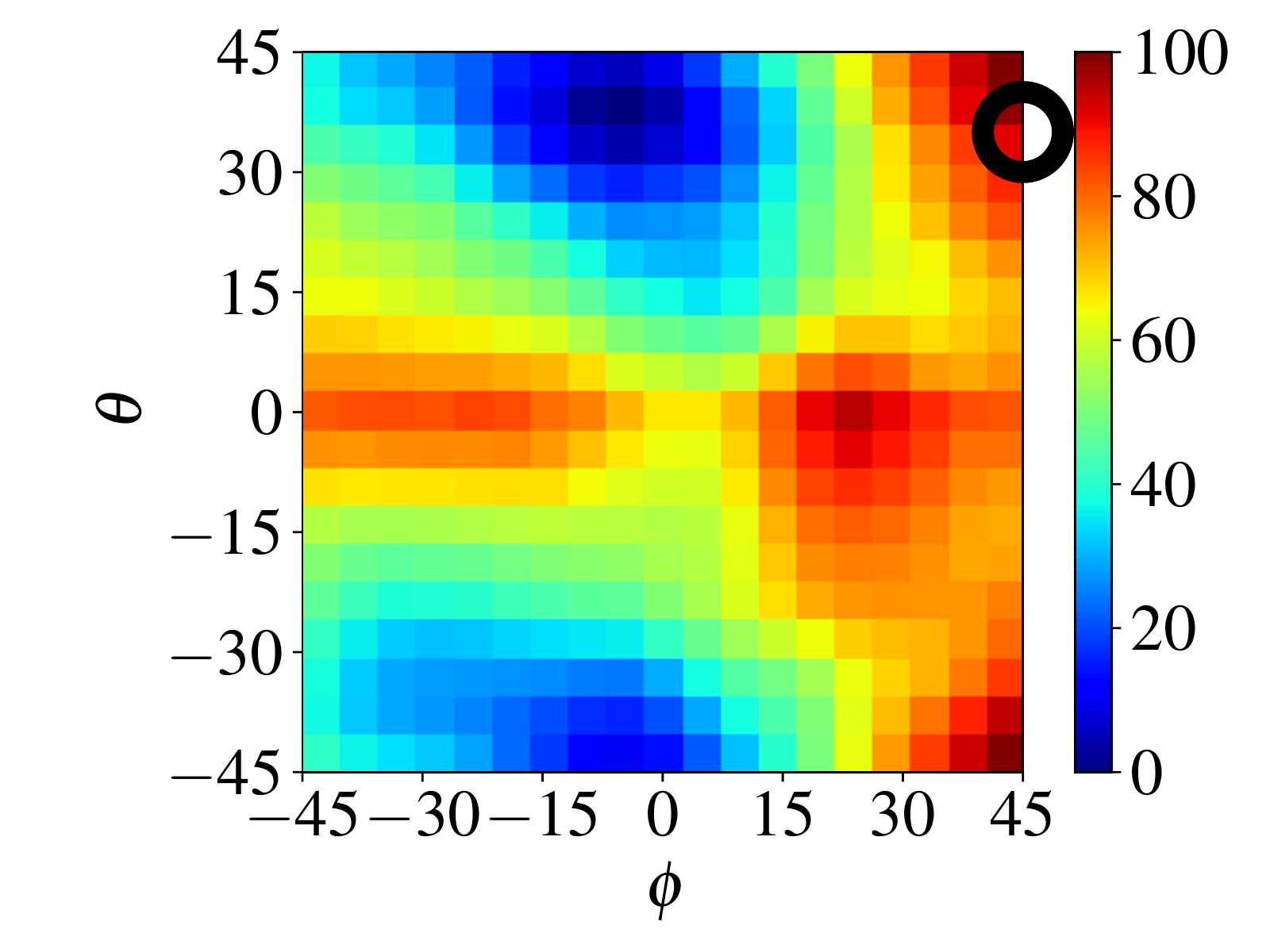}
    \end{subfigure}
    \begin{subfigure}[c]{0.33\linewidth}
        \centering
        \includegraphics[width=\textwidth]{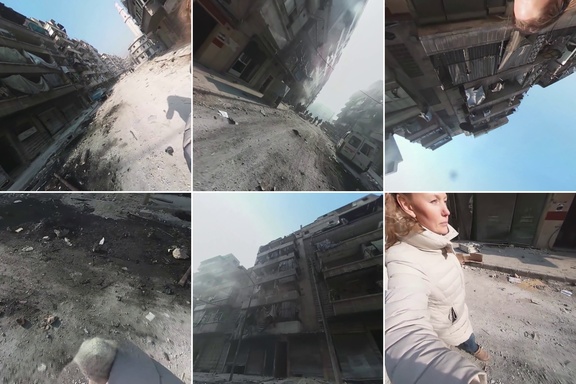}
    \end{subfigure}
    \begin{subfigure}[c]{0.33\linewidth}
        \centering
        \includegraphics[width=\textwidth]{./Figures/qualitative/VI4sPuwvOVk-seg101-min-1-enc.jpg}
    \end{subfigure}
    \\
    \begin{subfigure}[c]{0.31\linewidth}
        \centering
        \includegraphics[trim={1.2cm 0 0 0},clip,width=\textwidth]{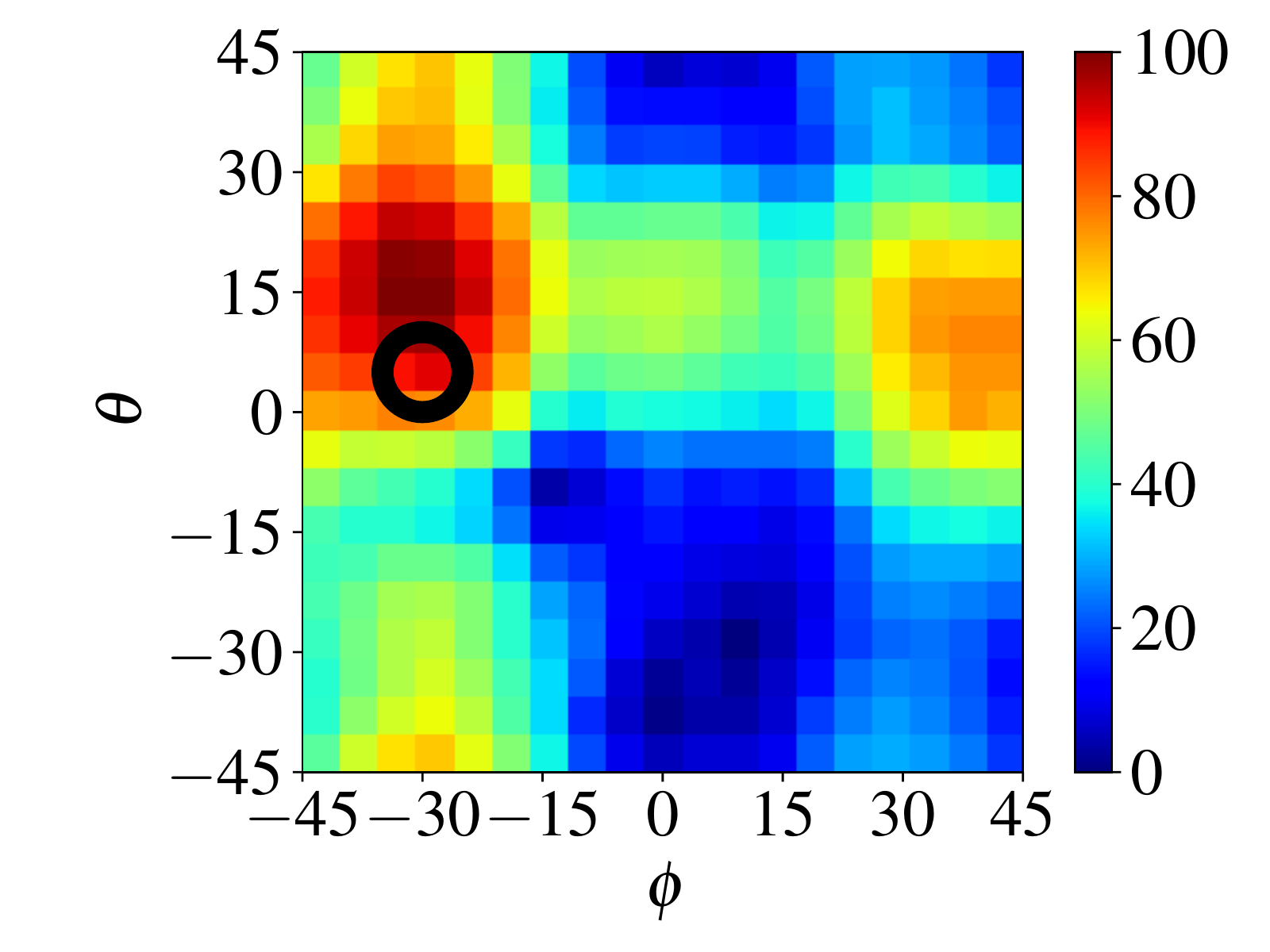}
    \end{subfigure}
    \begin{subfigure}[c]{0.33\linewidth}
        \centering
        \includegraphics[width=\textwidth]{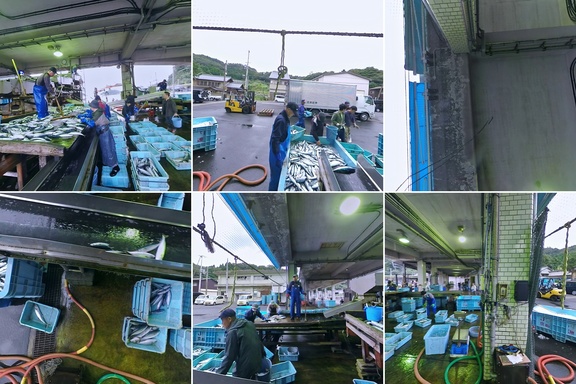}
    \end{subfigure}
    \begin{subfigure}[c]{0.33\linewidth}
        \centering
        \includegraphics[width=\textwidth]{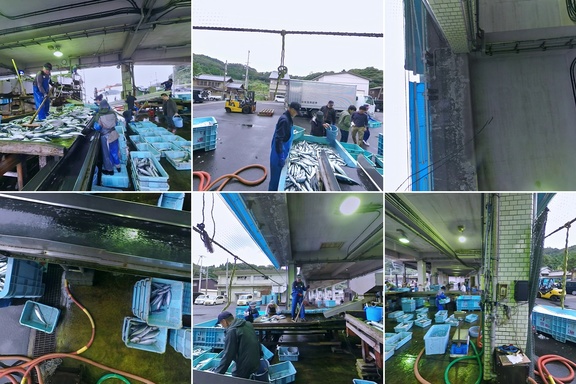}
    \end{subfigure}
    \\
    \begin{subfigure}[c]{0.31\linewidth}
        \centering
        \includegraphics[trim={1.2cm 0 0 0},clip,width=\textwidth]{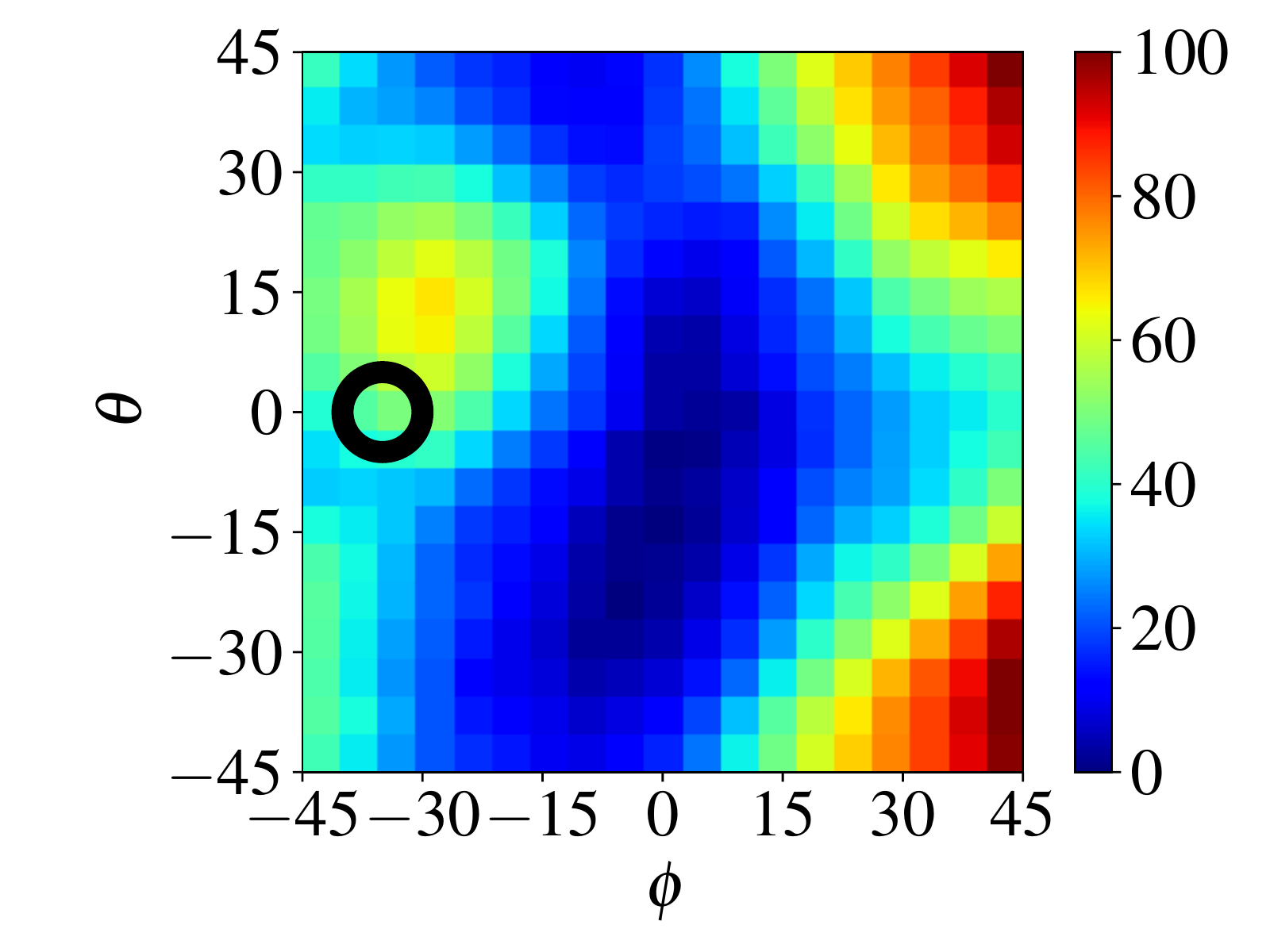}
    \end{subfigure}
    \begin{subfigure}[c]{0.33\linewidth}
        \centering
        \includegraphics[width=\textwidth]{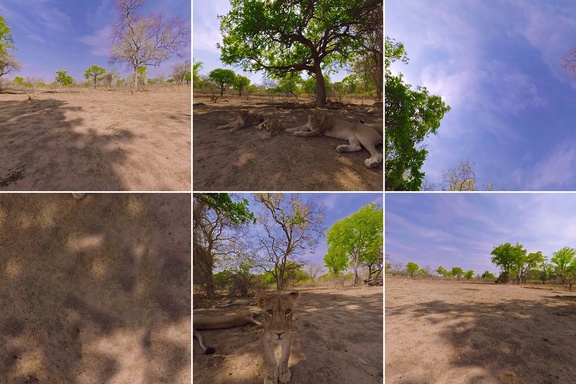}
    \end{subfigure}
    \begin{subfigure}[c]{0.33\linewidth}
        \centering
        \includegraphics[width=\textwidth]{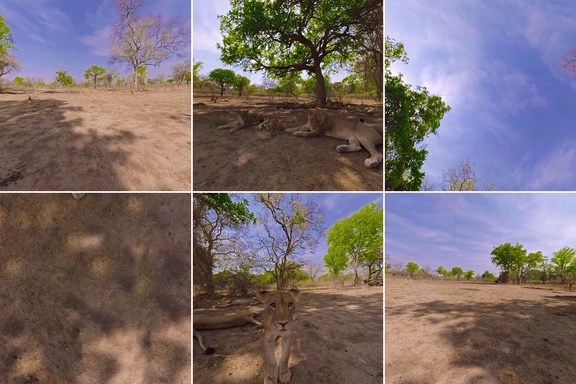}
    \end{subfigure}
    \\
    \caption{
        Qualitative examples.
        The heatmap shows the normalized reduction, and the overlaid circle shows our predicted result.
        The two images are the first and last frame of the clip rendered in the predicted orientation.
        Last row shows a failure example.
        Best viewed in color.
    }
    \label{fig:qualitative}
\end{figure}

Fig.~\ref{fig:qualitative} shows example prediction results.
Our approach performs well despite the diversity in the video content and recording situation.
The complexity in the content would make it hard to design a simple rule-based method to predict $\Omega^{min}$
(such as analyzing the continuity in Fig.~\ref{fig:concept_example}); a learning based method is necessary.
The last row shows a failure case of our method,
where the distribution of video size is multimodal,
and the model selects the suboptimal mode.

We next examine whether the model can be transferred across video formats,
e.g.~can the model trained on H264 videos improve the compression rate of HEVC videos?
Table~\ref{tab:transcode} shows the results.  
Overall, the results show our approach is capable of generalizing across video formats given common features.  We find that the model trained on H264 is less transferable,
while the models trained on HEVC and VP9 perform fairly well on H264.
In particular, the model trained on HEVC performs the best across all formats.
The reasons are twofold.
First, the models trained on HEVC and VP9 focus on the appearance feature which is common across all formats.
Second, the models trained on H264 suffer more from overfitting because the distribution of $\Omega^{min}$ is more concentrated.

\begin{table}[t]
    \small
    \center
    \begin{tabular}{cccccc}
    \toprule
        \multicolumn{2}{c}{H264} & \multicolumn{2}{c}{HEVC} & \multicolumn{2}{c}{VP9}\\
        \cmidrule(lr){1-2} \cmidrule(lr){3-4} \cmidrule(lr){5-6}
        HEVC & VP9 & H264 & VP9 & H264 & HEVC\\
        \midrule
        70.82 & 78.17 & 85.79 & 84.61 & 83.19 & 75.16 \\
    \bottomrule
    \end{tabular}
    \caption{
        Size reduction of our approach. Top row indicates training source, second row is test sources.
    }
    \label{tab:transcode}
\end{table}

\begin{figure}[t]
    \center
    \begin{subfigure}[c]{0.48\linewidth}
        \centering
        \includegraphics[trim={0.4cm 0 0 0},clip,width=.98\textwidth]{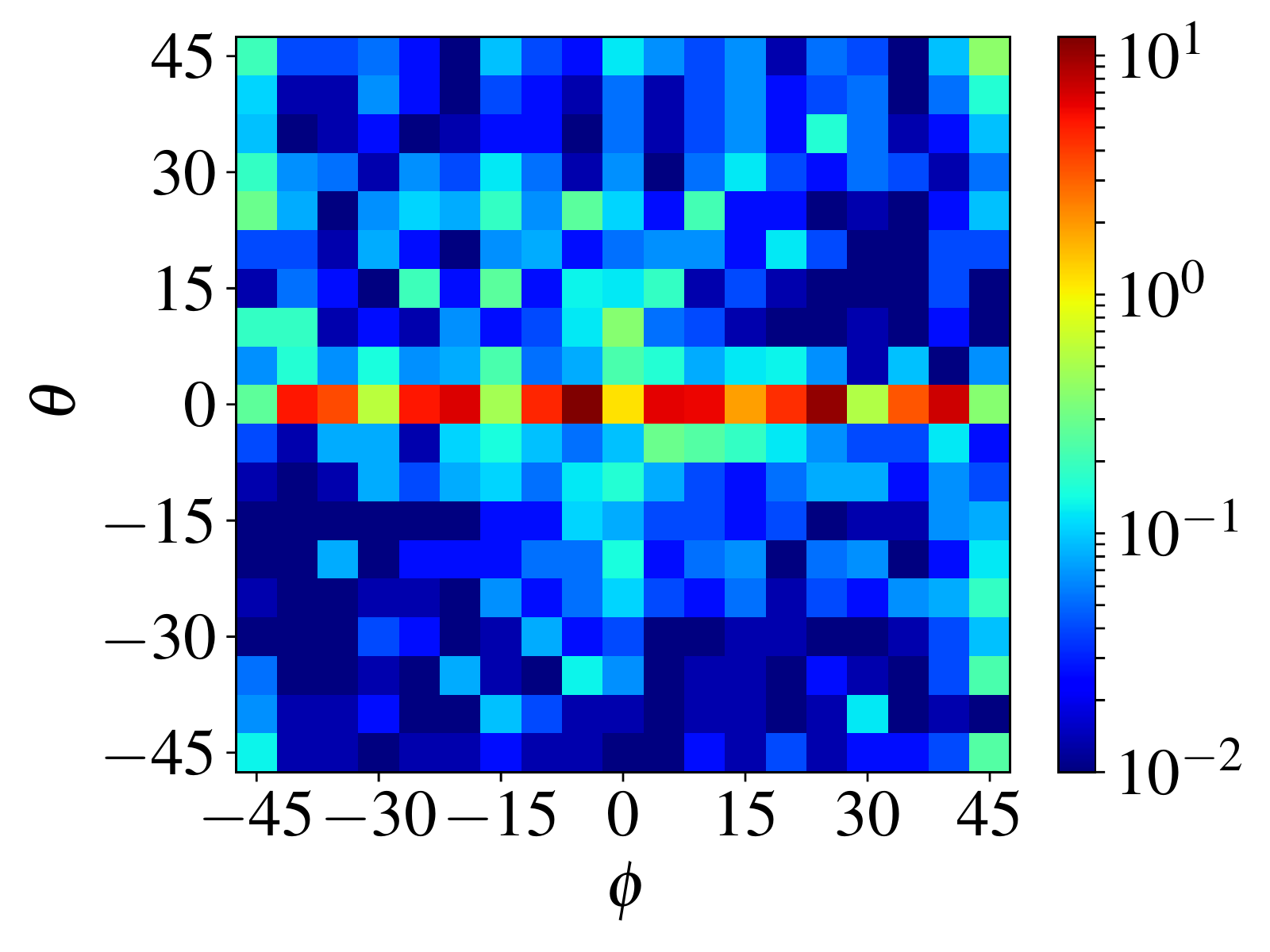}
        \caption{
            Predicted $\Omega^{min}$ (H264).
        }
    \end{subfigure}
    ~
    \begin{subfigure}[c]{0.48\linewidth}
        \centering
        \includegraphics[trim={0.4cm 0 0 0},clip,width=.98\textwidth]{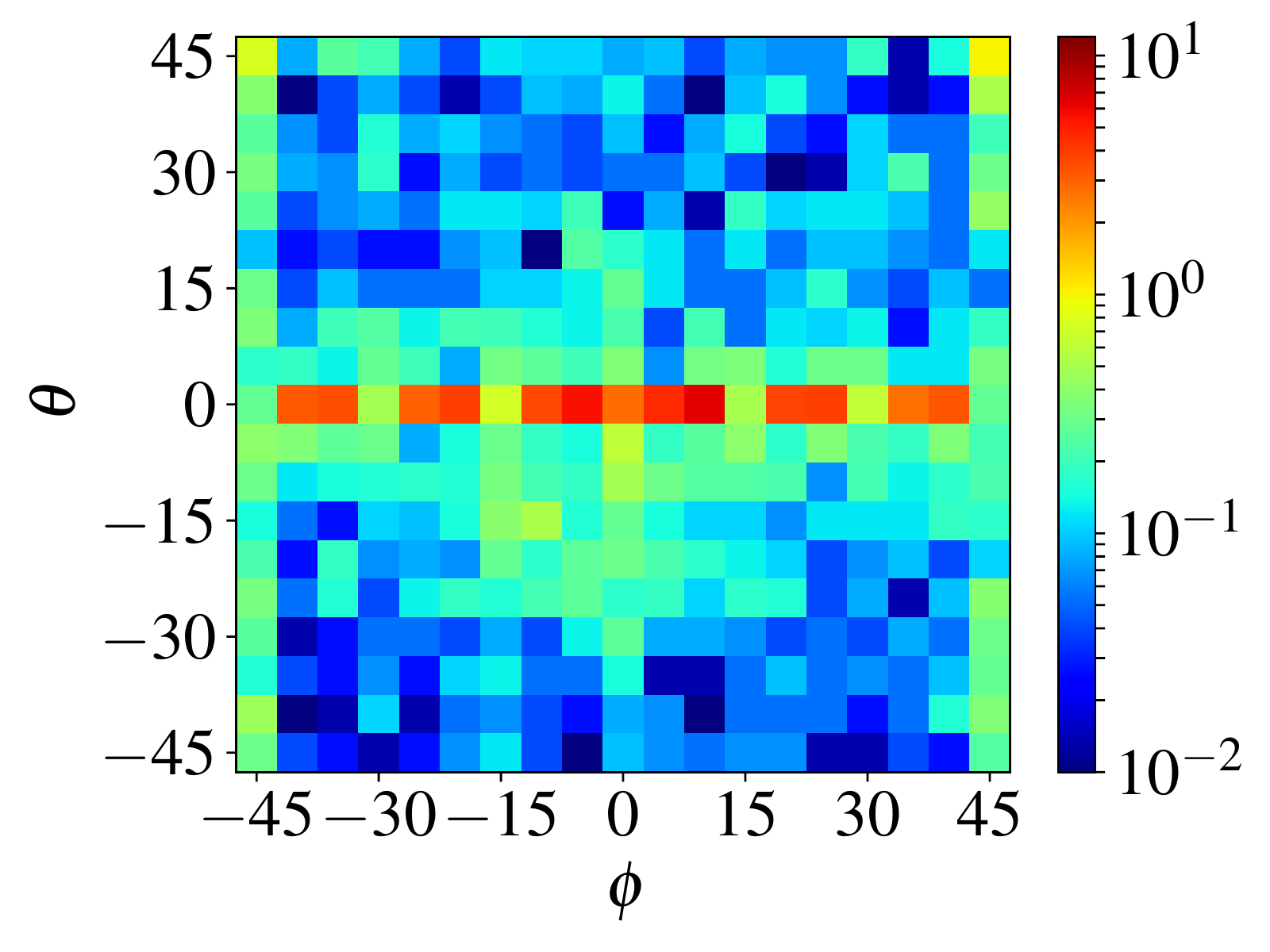}
        \caption{
            Real $\Omega^{min}$ of H264.
        }
    \end{subfigure}
    \begin{subfigure}[c]{0.48\linewidth}
        \centering
        \includegraphics[trim={0.4cm 0 0 0},clip,width=.98\textwidth]{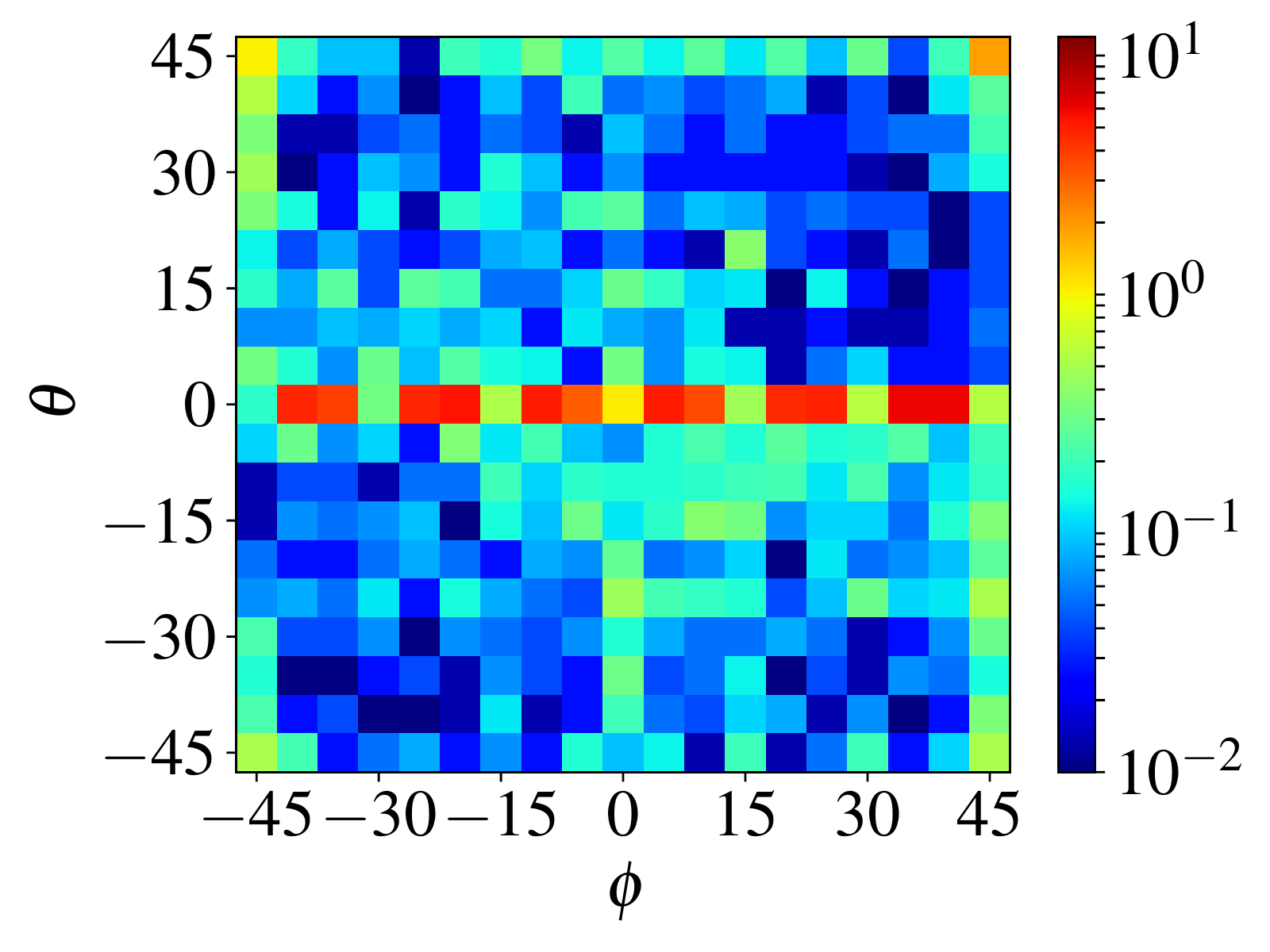}
        \caption{
            Predicted $\Omega^{min}$ (HEVC).
        }
    \end{subfigure}
    ~
    \begin{subfigure}[c]{0.48\linewidth}
        \centering
        \includegraphics[trim={0.4cm 0 0 0},clip,width=.98\textwidth]{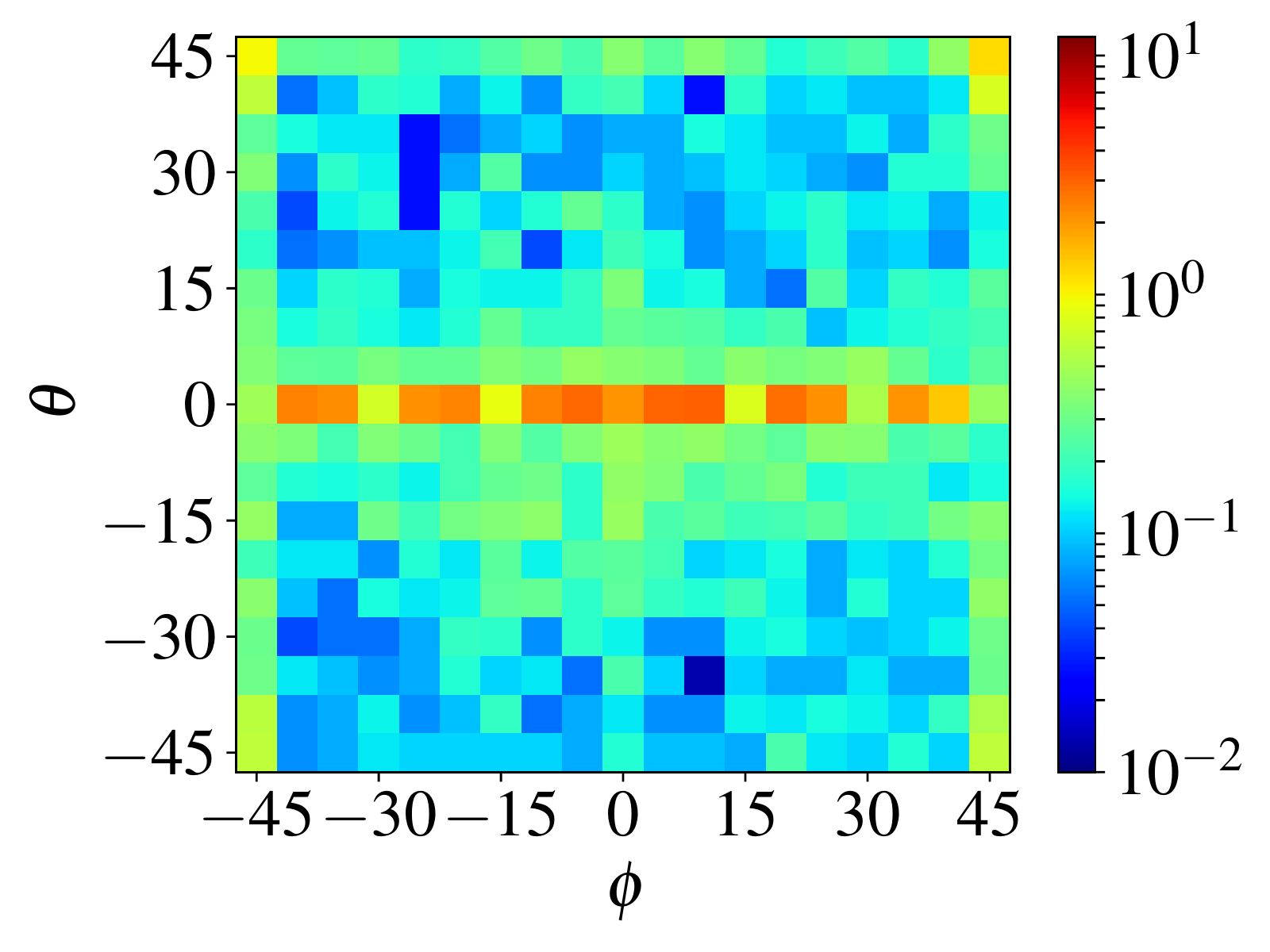}
        \caption{
            Real $\Omega^{min}$ of HEVC.
        }
    \end{subfigure}
    \caption{
        Distribution of $\Omega^{min}$ (\%).
        Predictions are on H264 videos with different training data.
    }
    \label{fig:prediction_distribution}
\end{figure}

The distribution of $\Omega^{min}$ provides further insight into the advantage of the model trained on HEVC.
See Fig.~\ref{fig:prediction_distribution}.
The predicted $\Omega^{min}$ tend to be more concentrated around $\theta{=}0$ than the real $\Omega^{min}$.
Because the distribution of $\Omega^{min}$ is more dispersed in HEVC,
so is the prediction of $\Omega^{min}$ by the model trained on HEVC.

\section{Conclusion}

This work studies how to improve $360\degree$ video compression by selecting a proper orientation for cubemap projection.
Our analysis across 3 popular codecs shows scope for reducing video sizes by up to $76\%$ through rotation,
with an average of more than $8\%$ over all videos.
We propose an approach that predicts the optimal orientation given the video in a single orientation.
It achieves $82\%$ the compression rate of the optimal orientation while requiring less than $0.3\%$ of the computation of a non-learned solution (fraction of a second vs.~1.5 hours per GOP).
Future work will explore how to combine orientation search and prediction to reach better compression rates under a computational budget.

%
\IEEEpeerreviewmaketitle

\appendices

\section{Compression Parameters}

We encode the videos using x264/x265/libvpx through FFMPEG.
The compression parameters of FFMPEG for each encoder are as follows.
\begin{itemize}[leftmargin=*,label=$\bullet$]
    \item x264 --- ``-preset medium -crf 0 -an''
    \item x265 --- ``-preset medium -x265-params lossless=1 -crf 0 -an''
    \item libvpx --- ``-speed 4 -cpu-used 4 -lossless 1 -qmin 0 -qmax 0 -an''
\end{itemize}
The ``-an'' option disables audio in the output bit-stream.
The ``-preset medium'' in x264/x264 and ``-speed 4 -cpu-used 4'' controls the encoding speed.
We use the default setting for x264/x265 which provides a reasonable balance between speed and compression rate.
Other options specify lossless compression for each encoder.
For the transform360 filter,
we use bicubic interpolation for pixel values.

\section{Rotational Symmetry}

We justify our design of restricting the rotation within $90\degree$.
We compute the correlation between cubemap size related by $90\degree$ rotation along either $\theta$ or $\phi$.
In order to do so,
we compute the size of cubemaps with $\{\pm60\degree, \pm75\degree, \pm90\degree\}$ rotation along either pitch or yaw and compare them with those with rotation within $[-45\degree, 45\degree]$.
The correlations are in Table~\ref{tab:rotation}.
We also show the correlations for $45\degree$ rotation for comparison.
The strong correlation clearly shows that the cubemap sizes are indeed symmetric to $90\degree$ rotation.

\begin{table}[h]
    \small
    \center
    \begin{tabular}{cccc}
    \toprule
    Encoders & H264 & HEVC & VP9\\
    \midrule
        $90\degree$ & 1.00  & 0.95 & 0.99\\
        $45\degree$ & 0.25  & 0.23 & 0.23 \\
    \bottomrule
    \end{tabular}
    \caption{
        Correlation between cubemap sizes related by $90\degree$ and $45\degree$ rotation.
    }
    \label{tab:rotation}
\end{table}

\section{Qualitative Examples}

\begin{figure*}[t]
    \vspace{-4pt}
    \center
    \hspace{0.19\linewidth}
    ~
    \begin{subfigure}[c]{0.38\linewidth}
        \centering
        \caption{Ours}
    \end{subfigure}
    ~~~
    \begin{subfigure}[c]{0.38\linewidth}
        \centering
        \caption{\textsc{Center} baseline}
    \end{subfigure}
    \\
    \vspace{-12pt}
    \InsertQualitative{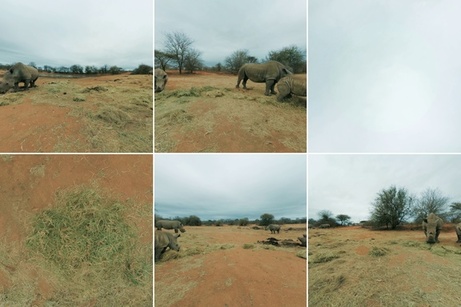}\\
    \vspace{-1pt}
    \InsertQualitative{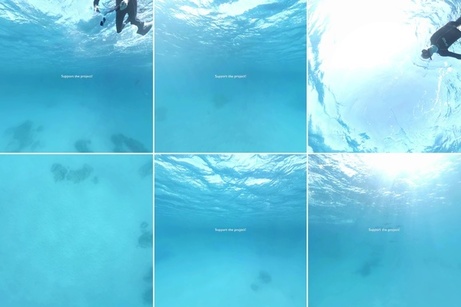}\\
    \vspace{-1pt}
    \InsertQualitative{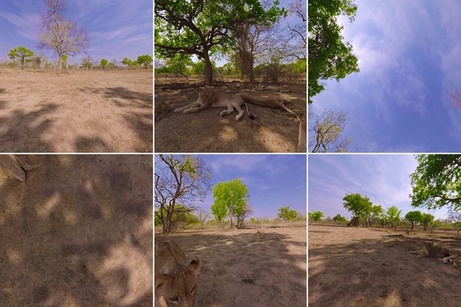}\\
    \vspace{-1pt}
    \InsertQualitative{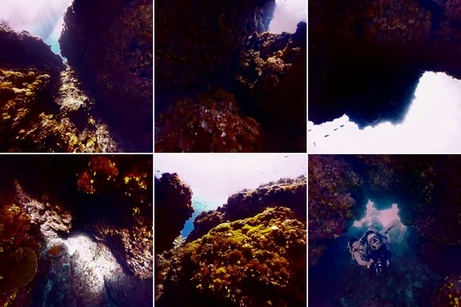}\\
    \vspace{-1pt}
    \InsertQualitative{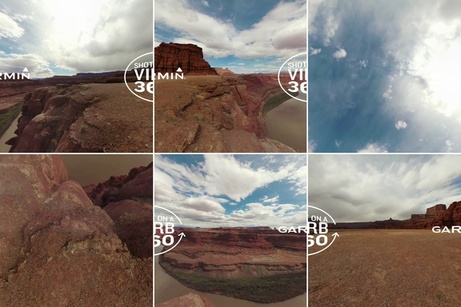}\\
    \vspace{-1pt}
    \InsertQualitative{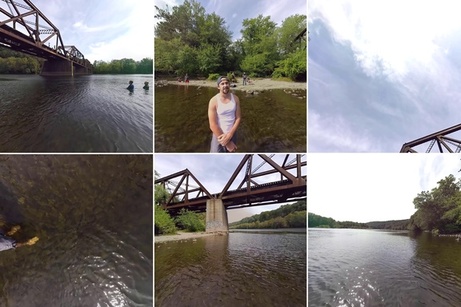}\\
    \vspace{-1pt}
    \InsertQualitative{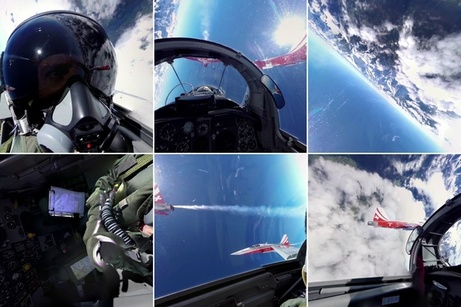}\\
    \vspace{-1pt}
    \InsertQualitative{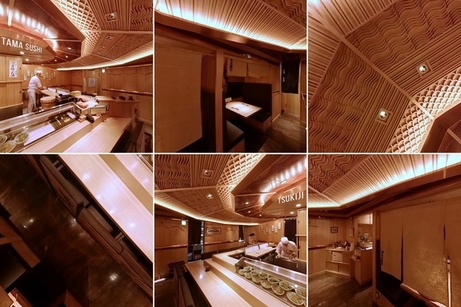}
    \\
    \vspace{-9pt}
    \caption{
        Qualitative results.
        Each row shows a clip.
        The first figure per row shows the size distribution.
        Black circle shows the predicted result, which is rendered in the second and third figures.
        The fourth and fifth figures show the \textsc{Center} baseline.
        The second and fourth figures show the first frame of the clip, and the third and fifth figures are the last frame.
    }
    \label{fig:qualitative2}
\end{figure*}

\begin{figure*}[t]
    \vspace{-8pt}
    \center
    \hspace{0.19\linewidth}
    ~
    \begin{subfigure}[c]{0.38\linewidth}
        \centering
        \caption{Ours}
    \end{subfigure}
    ~~~
    \begin{subfigure}[c]{0.38\linewidth}
        \centering
        \caption{\textsc{Center} baseline}
    \end{subfigure}
    \\
    \vspace{-12pt}
    \InsertQualitative{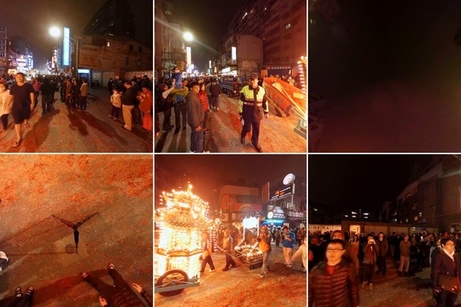}
    \vspace{-4pt}
    \InsertQualitative{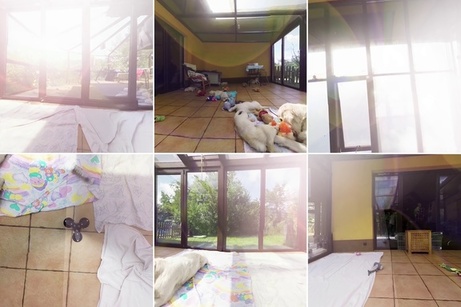}
    \vspace{-4pt}
    \InsertQualitative{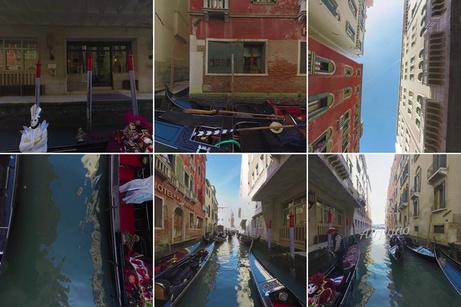}
    \vspace{-4pt}
    \InsertQualitative{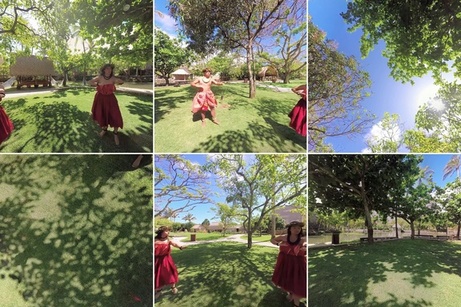}
    \\
    \vspace{-8pt}
    \caption{
        Qualitative results (cont.).
    }
    \label{fig:qualitative3}
    \vspace{-6pt}
\end{figure*}

In this section, we show more qualitative examples similar to Fig.~\ref{fig:qualitative} in the main paper.
See Fig.~\ref{fig:qualitative2} and \ref{fig:qualitative3}.
We can see that even small objects can affect the compression rate,
such as the rhinos in the first example and the diver in the second example.
The pattern doesn't even have to correspond to a real object like the blank region in the fourth example or the logo in the fifth example.
The fifth example also shows how the file size is affected by multiple factors jointly.
The distribution would be symmetric with respect to $\theta{=}0$ if the file size only depended on the logo,
but the sky and cloud lead to the additional mode at the top middle.
We also see that the continuity of foreground objects is not the only factor that matters from the sixth and seventh example;
the person in the sixth example and the pilot in the seventh example lie on the face boundary in the optimal orientation.
The result suggests that heuristics based on object location, either automatic or manual, do not solve the problem.
The eighth example shows that the compression would be more efficient if the motions fall in the same face even if it does not introduce discontinuity in motion.

\section{Failure Cases}

\begin{figure*}[t]
    \center
    \hspace{0.19\linewidth}
    ~
    \begin{subfigure}[c]{0.38\linewidth}
        \centering
        \caption{Ours}
    \end{subfigure}
    ~~~
    \begin{subfigure}[c]{0.38\linewidth}
        \centering
        \caption{\textsc{Center} baseline}
    \end{subfigure}
    \\
    \vspace{-12pt}
    \InsertFailure{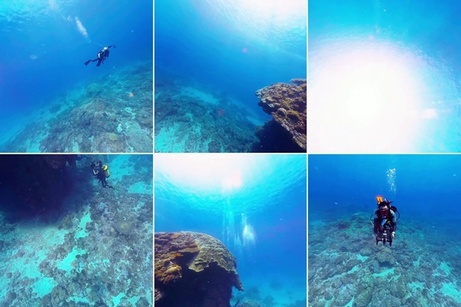}
    \vspace{-4pt}
    \InsertFailure{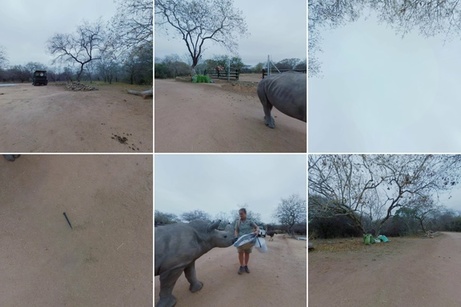}
    \vspace{-4pt}
    \InsertFailure{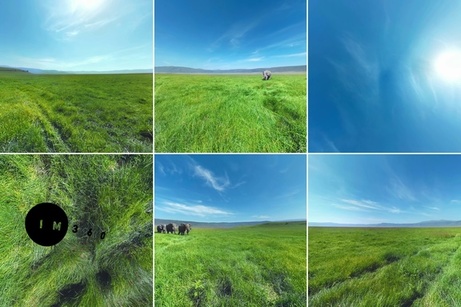}
    \vspace{-4pt}
    \InsertFailure{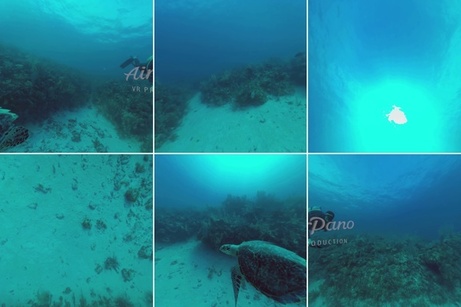}
    \\
    \vspace{-6pt}
    \caption{
        Failure cases.
        Blue circle shows true minimum and is rendered in the fourth and fifth figures.
        The two figures are the first and last frame of the clip.
    }
    \label{fig:failures}
    \vspace{-9pt}
\end{figure*}

In this section, we show failure examples similar to Fig.~\ref{fig:qualitative} in the main paper.
See Fig.~\ref{fig:failures}.
In the first example,
we can see the best compression rate occurs when the coral is continuous,
while our method fails because it decides to keep the sun light (round white pattern) continuous.
In the second and third example,
the video size tends to be smaller when the horizon falls on the face diagonal,
possibly because it is more friendly for intra-prediction in compression.
Our method doesn't learn this tendency,
so it fails to predict the optimal $\theta$ and only predicts the correct $\phi$.

\clearpage

\ifCLASSOPTIONcaptionsoff
  \newpage
\fi



%

\bibliographystyle{IEEEtran}
\bibliography{isomer360}

\begin{thebibliography}{10}
\providecommand{\url}[1]{#1}
\csname url@samestyle\endcsname
\providecommand{\newblock}{\relax}
\providecommand{\bibinfo}[2]{#2}
\providecommand{\BIBentrySTDinterwordspacing}{\spaceskip=0pt\relax}
\providecommand{\BIBentryALTinterwordstretchfactor}{4}
\providecommand{\BIBentryALTinterwordspacing}{\spaceskip=\fontdimen2\font plus
\BIBentryALTinterwordstretchfactor\fontdimen3\font minus
  \fontdimen4\font\relax}
\providecommand{\BIBforeignlanguage}[2]{{%
\expandafter\ifx\csname l@#1\endcsname\relax
\typeout{** WARNING: IEEEtran.bst: No hyphenation pattern has been}%
\typeout{** loaded for the language `#1'. Using the pattern for}%
\typeout{** the default language instead.}%
\else
\language=\csname l@#1\endcsname
\fi
#2}}
\providecommand{\BIBdecl}{\relax}
\BIBdecl

\bibitem{360camera}
V.~Ukonaho, ``Global 360 camera sales forecast by segment: 2016 to 2022,''
  \url{https://www.strategyanalytics.com/access-services/devices/mobile-phones/emerging-devices/market-data/report-detail/global-360-camera-sales-forecast-by-segment-2016-to-2022},
  March 2017.

\bibitem{fb360videostatistics}
B.~Ayrey and C.~Wong, ``Introducing facebook 360 for gear vr,''
  \url{https://newsroom.fb.com/news/2017/03/introducing-facebook-360-for-gear-vr/},
  March 2017.

\bibitem{kasahara2015first}
S.~Kasahara, S.~Nagai, and J.~Rekimoto, ``First person omnidirectional video:
  System design and implications for immersive experience,'' in \emph{ACM TVX},
  2015.

\bibitem{kopf2016tog}
J.~Kopf, ``360$\degree$ video stabilization,'' \emph{ACM Transactions on
  Graphics (TOG)}, vol.~35, no.~6, p. 195, 2016.

\bibitem{kamali2011stabilizing}
M.~Kamali, A.~Banno, J.-C. Bazin, I.~S. Kweon, and K.~Ikeuchi, ``Stabilizing
  omnidirectional videos using 3d structure and spherical image warping,'' in
  \emph{IAPR MVA}, 2011.

\bibitem{su2016accv}
Y.-C. Su, D.~Jayaraman, and K.~Grauman, ``Pano2vid: Automatic cinematography
  for watching $360\degree$ videos,'' in \emph{ACCV}, 2016.

\bibitem{su2017cvpr}
Y.-C. Su and K.~Grauman, ``Making $360\degree$ video watchable in 2d: Learning
  videography for click free viewing.'' in \emph{CVPR}, 2017.

\bibitem{hu2017deep}
H.-N. Hu, Y.-C. Lin, M.-Y. Liu, H.-T. Cheng, Y.-J. Chang, and M.~Sun, ``Deep
  360 pilot: Learning a deep agent for piloting through $360\degree$ sports
  video,'' in \emph{CVPR}, 2017.

\bibitem{lai2017semantic}
W.-S. Lai, Y.~Huang, N.~Joshi, C.~Buehler, M.-H. Yang, and S.~B. Kang,
  ``Semantic-driven generation of hyperlapse from 360\degree~video,''
  \emph{IEEE Transactions on Visualization and Computer Graphics}, vol.~PP,
  no.~99, pp. 1--1, 2017.

\bibitem{khasanova2017graph}
R.~Khasanova and P.~Frossard, ``Graph-based classification of omnidirectional
  images,'' \emph{arXiv preprint arXiv:1707.08301}, 2017.

\bibitem{cohen2017convolutional}
T.~Cohen, M.~Geiger, and M.~Welling, ``Convolutional networks for spherical
  signals,'' \emph{arXiv preprint arXiv:1709.04893}, 2017.

\bibitem{su2017nips}
Y.-C. Su and K.~Grauman, ``Flat2sphere: Learning spherical convolution for fast
  features from $360\degree$ imagery,'' in \emph{NIPS}, 2017.

\bibitem{fb2015cubemap}
E.~Kuzyakov and D.~Pio, ``{Under the hood: Building 360 video},''
  \url{https://code.facebook.com/posts/1638767863078802/under-the-hood-building-360-video/},
  October 2015.

\bibitem{google2017eac}
C.~Brown, ``{Bringing pixels front and center in VR video},''
  \url{https://www.blog.google/products/google-vr/bringing-pixels-front-and-center-vr-video/},
  March 2017.

\bibitem{mpeg120}
{Moving Picture Experts Group}, ``Point cloud compression – mpeg evaluates
  responses to call for proposal and kicks off its technical work [press
  release],'' \url{https://mpeg.chiariglione.org/meetings/120}, October 2017.

\bibitem{fb2016compressionrate}
E.~Kuzyakov and D.~Pio, ``{Next-generation video encoding techniques for 360
  video and VR},''
  \url{https://code.facebook.com/posts/1126354007399553/next-generation-video-encoding-techniques-for-360-video-and-vr/},
  January 2016.

\bibitem{omaf2017wd}
B.~Choi, Y.-K. Wang, and M.~M. Hannuksela, ``Wd on iso/iec 23000-20
  omnidirectional media application format,'' ISO/IEC JTC1/SC29/WG11, 2017.

\bibitem{hansen2007scale}
P.~Hansen, P.~Corke, W.~Boles, and K.~Daniilidis, ``Scale-invariant features on
  the sphere,'' in \emph{ICCV}, 2007.

\bibitem{cfe2017JVET}
M.~Wien, V.~Baroncini, J.~Boyce, A.~Segall, and T.~Suzuki, ``Joint call for
  evidence on video compression with capability beyond hevc,'' JVET-F1002,
  2017.

\bibitem{JVET-G0023}
M.~Coban, G.~V. der Auwera, and M.~Karczewicz, ``Qualcomm’s response to joint
  cfe in 360-degree video category,'' JVET-G0023, 2017.

\bibitem{JVET-G0024}
P.~Hanhart, X.~Xiu, F.~Duanmu, Y.~He, and Y.~Ye, ``Interdigital’s response to
  the 360° video category in joint call for evidence on video compression with
  capability beyond hevc,'' JVET-G0024, 2017.

\bibitem{JVET-G0025}
E.~Alshina, K.~Choi, V.~Zakharchenko, S.~N. Akula, A.~Dsouza, C.~Pujara, K.~K.
  Ramkumaar, and A.~Singh, ``Samsung's response to joint cfe on video
  compression with capability beyond hevc (360° category),'' JVET-G0025, 2017.

\bibitem{JVET-G0026}
A.~Gabriel and E.~Thomas, ``Polyphase subsampling applied to 360-degree video
  sequences in the context of the joint call for evidence on video
  compression,'' JVET-G0026, 2017.

\bibitem{sanchez2015panohevc}
Y.~Sánchez, R.~Skupin, and T.~Schierl, ``Compressed domain video processing
  for tile based panoramic streaming using hevc,'' in \emph{ICIP}, 2015.

\bibitem{sreedhar2016adaptive}
K.~K. Sreedhar, A.~Aminlou, M.~M. Hannuksela, and M.~Gabbouj,
  ``Viewport-adaptive encoding and streaming of 360-degree video for virtual
  reality applications,'' in \emph{IEEE ISM}, 2016.

\bibitem{snyder1987map}
J.~P. Snyder, \emph{Map projections--A working manual}.\hskip 1em plus 0.5em
  minus 0.4em\relax US Government Printing Office, 1987, vol. 1395.

\bibitem{zelnik2005squaring}
L.~Zelnik-Manor, G.~Peters, and P.~Perona, ``Squaring the circle in
  panoramas,'' in \emph{ICCV}, 2005.

\bibitem{kim-iccv2017}
Y.~Kim, C.-R. Lee, D.-Y. Cho, Y.~Kwon, H.-J. Choi, and K.-J. Yoon, ``Automatic
  content-award projection for 360$\degree$ videos,'' in \emph{ICCV}, 2017.

\bibitem{chang-iccv2013}
C.-H. Chang, M.-C. Hu, W.-H. Cheng, and Y.-Y. Chuang, ``Rectangling
  stereographic projection for wide-angle image visualization,'' in
  \emph{ICCV}, 2013.

\bibitem{adeel2017rsp}
D.~N. Adeel~Abbas, ``A novel projection for omni-directional video,'' in
  \emph{Proc.SPIE 10396}, 2017.

\bibitem{santurkar2017generative}
S.~Santurkar, D.~Budden, and N.~Shavit, ``Generative compression,'' \emph{arXiv
  preprint arXiv:1703.01467}, 2017.

\bibitem{rippel2017real}
O.~Rippel and L.~Bourdev, ``Real-time adaptive image compression,'' in
  \emph{ICML}, 2017.

\bibitem{toderici2015variable}
G.~Toderici, S.~M. O'Malley, S.~J. Hwang, D.~Vincent, D.~Minnen, S.~Baluja,
  M.~Covell, and R.~Sukthankar, ``Variable rate image compression with
  recurrent neural networks,'' in \emph{ICLR}, 2016.

\bibitem{toderici2016full}
G.~Toderici, D.~Vincent, N.~Johnston, S.~J. Hwang, D.~Minnen, J.~Shor, and
  M.~Covell, ``Full resolution image compression with recurrent neural
  networks,'' in \emph{CVPR}, 2017.

\bibitem{johnston2017improved}
N.~Johnston, D.~Vincent, D.~Minnen, M.~Covell, S.~Singh, T.~Chinen, S.~J.
  Hwang, J.~Shor, and G.~Toderici, ``Improved lossy image compression with
  priming and spatially adaptive bit rates for recurrent networks,''
  \emph{arXiv preprint arXiv:1703.10114}, 2017.

\bibitem{li2017learning}
M.~Li, W.~Zuo, S.~Gu, D.~Zhao, and D.~Zhang, ``Learning convolutional networks
  for content-weighted image compression,'' \emph{arXiv preprint
  arXiv:1703.10553}, 2017.

\bibitem{videocompression}
K.~R. Rao, D.~N. Kim, and J.~J. Hwang, \emph{Video Coding Standards: AVS China,
  H.264/MPEG-4 PART 10, HEVC, VP6, DIRAC and VC-1}.\hskip 1em plus 0.5em minus
  0.4em\relax Springer Netherlands, 2014.

\bibitem{bluray}
{Blu-ray Disc Association}, ``White paper blu-ray disc read-only format coding
  constraints on hevc video streams for bd-rom version 3.0,'' June 2015.

\bibitem{achanta2012slic}
R.~Achanta, A.~Shaji, K.~Smith, A.~Lucchi, P.~Fua, and S.~S{\"u}sstrunk, ``Slic
  superpixels compared to state-of-the-art superpixel methods,'' \emph{IEEE
  transactions on pattern analysis and machine intelligence}, vol.~34, no.~11,
  pp. 2274--2282, 2012.

\bibitem{simonyan2014very}
K.~Simonyan and A.~Zisserman, ``Very deep convolutional networks for
  large-scale image recognition,'' in \emph{ICLR}, 2015.

\bibitem{long2015fully}
J.~Long, E.~Shelhamer, and T.~Darrell, ``Fully convolutional networks for
  semantic segmentation,'' in \emph{CVPR}, 2015.

\bibitem{glorot2010initialization}
X.~Glorot and Y.~Bengio, ``Understanding the difficulty of training deep
  feedforward neural networks,'' in \emph{AISTATS}, 2010.

\bibitem{kingma2014adam}
D.~Kingma and J.~Ba, ``Adam: A method for stochastic optimization,'' in
  \emph{ICLR}, 2015.

\end{thebibliography}

\end{document}